\documentclass{article}

\usepackage{microtype}
\usepackage{graphicx}
\usepackage{subfigure}
\usepackage{booktabs} 

\usepackage{hyperref}

\usepackage[accepted]{icml2024}

\usepackage{amsmath}
\usepackage{amssymb}
\usepackage{mathtools}
\usepackage{amsthm}

\usepackage[capitalize,noabbrev]{cleveref}

\theoremstyle{plain}

\theoremstyle{definition}

\theoremstyle{remark}

\usepackage[textsize=tiny]{todonotes}

\icmltitlerunning{Think Before You Act: Decision Transformers with Working Memory}

\usepackage{amsfonts}
\usepackage{subcaption}
\usepackage{algorithm}
\usepackage{algorithmic}
\usepackage{amsmath}
\usepackage{graphicx}
\usepackage{mathtools}
\usepackage{bbm}
\usepackage{multirow}
\usepackage{bm}
\usepackage{wrapfig}
\usepackage{enumitem}
\setlist{nolistsep}
\usepackage{adjustbox}
\usepackage{lineno}
\usepackage{xspace}

\definecolor{jikun}{HTML}{FA7F6F}
\definecolor{romain}{HTML}{4995c6}

\definecolor{r1}{HTML}{80A6E2}
\definecolor{r2}{HTML}{70C17F}
\definecolor{r3}{HTML}{F46F43}
\definecolor{r4}{HTML}{FBDD85}

\newcommand{\name}{DT-Mem\xspace}
\newcommand{\fullname}{\textbf{D}ecision \textbf{T}ransformers with \textbf{M}emory\xspace}

\begin{document}

\twocolumn[
\icmltitle{Think Before You Act: Decision Transformers with Working Memory}



\icmlsetsymbol{equal}{*}

\begin{icmlauthorlist}
\icmlauthor{Jikun Kang}{mcgill,mila}
\icmlauthor{Romain Laroche}{equal}
\icmlauthor{Xingdi Yuan}{equal,msr}
\icmlauthor{Adam Trischler}{equal}
\icmlauthor{Xue Liu}{equal,mcgill,mila}
\icmlauthor{Jie Fu}{mila}
\end{icmlauthorlist}

\icmlaffiliation{mcgill}{Department of Computer Science, McGill University, Montr\'{e}al, Canada}
\icmlaffiliation{mila}{Mila - Qu\'ebec AI Institute, Montr\'{e}al, Canada}
\icmlaffiliation{msr}{Microsoft Research, Montr\'{e}al, Canada}

\icmlcorrespondingauthor{Jikun Kang}{jikun.kang@mail.mcgill.ca}
\icmlcorrespondingauthor{Jie Fu}{jie.fu@polymtl.ca}

\icmlkeywords{Machine Learning, ICML}

\vskip 0.3in
]



\printAffiliationsAndNotice{*Equal advising}  

\begin{abstract}
Decision Transformer-based decision-making agents have shown the ability to generalize across multiple tasks. However, their performance relies on massive data and computation. We argue that this inefficiency stems from the forgetting phenomenon, in which a model memorizes its behaviors in parameters throughout training. As a result, training on a new task may deteriorate the model's performance on previous tasks. In contrast to LLMs' implicit memory mechanism, the human brain utilizes distributed memory storage, which helps manage and organize multiple skills efficiently, mitigating the forgetting phenomenon. Inspired by this, we propose a working memory module to store, blend, and retrieve information for different downstream tasks. Evaluation results show that the proposed method improves training efficiency and generalization in Atari games and Meta-World object manipulation tasks. Moreover, we demonstrate that memory fine-tuning further enhances the adaptability of the proposed architecture.
\footnote{We open source the code at \url{https://github.com/luciferkonn/DT_Mem}.}
\end{abstract}

\section{Introduction}

\begin{figure}[htbp]
    \includegraphics[width=0.48\textwidth]{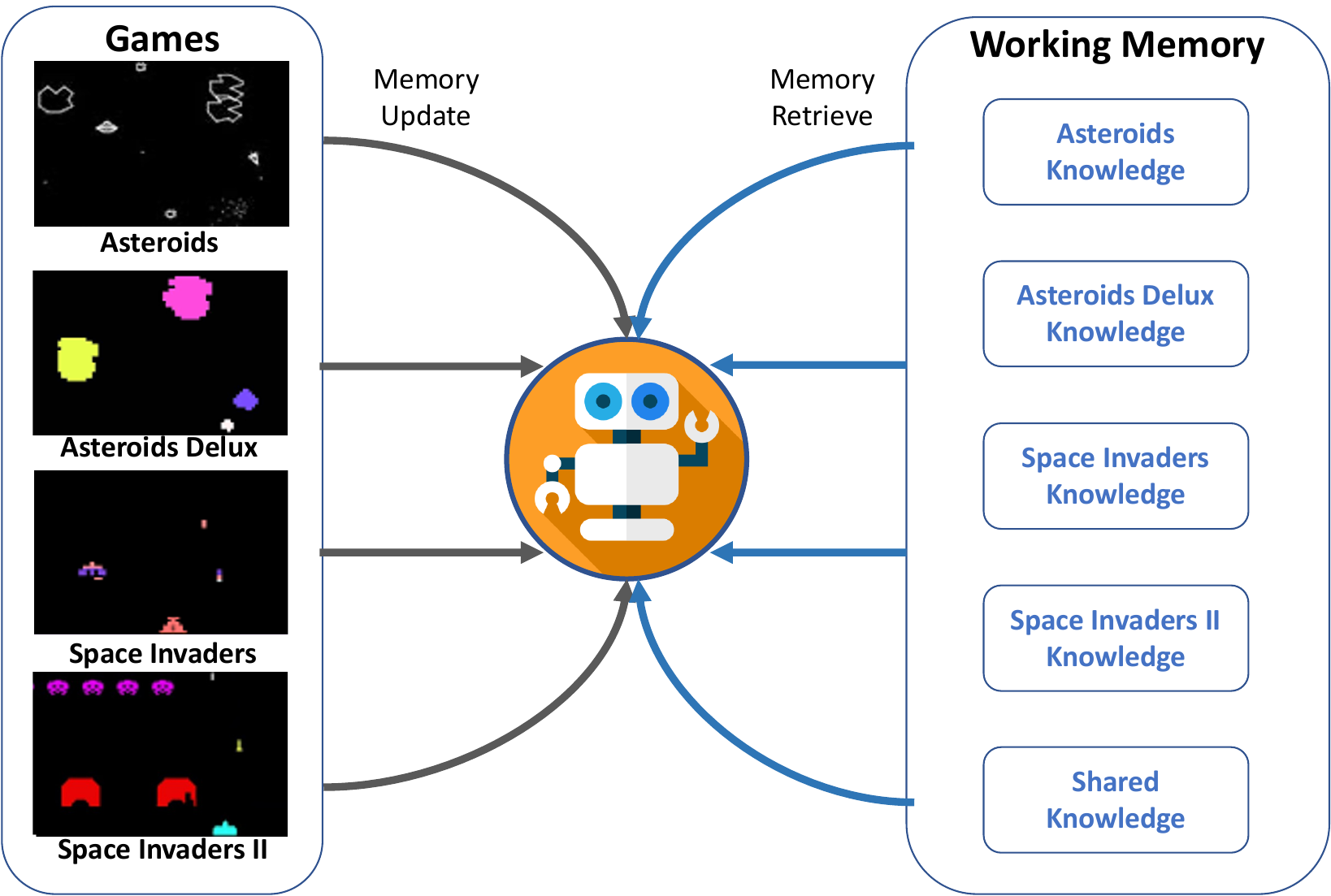}
    \caption{Illustrating how a robot can use its memory to guide its playing strategy.}
    \label{figs:intro}
\end{figure}

Recently, with the tremendous success of decoder-only Transformer models \citep{DBLP:journals/corr/abs-2005-14165, DBLP:journals/corr/abs-2303-08774, DBLP:conf/iclr/DosovitskiyB0WZ21, DBLP:journals/corr/abs-2302-13971}, an increasing number of researchers have focused on decoder-only Transformer-based decision-making agents. 
As shown with GPT-3 \citep{DBLP:journals/corr/abs-2005-14165} and follow-up work~\cite{kaplan2020scaling,DBLP:conf/icml/ClarkCGMPHDHCB022}, the generalization of these LLMs depends significantly on the model size, \textit{i.e.} the number of parameters.
This is partly because neural network parameters act as implicit memory \citep{DBLP:conf/iclr/NeyshaburLBLS19}, enabling models to ``memorize'' a huge amount of training data by fitting these parameters.
However, relying purely on scale has practical and ethical limits: there are economic and ecological costs, and it reduces accessibility
To address some limits of the implicit, parameter-based memory of large models, we take inspiration from the concept of ``working memory'' \citep{baddeley2003working, cowan2008differences} to \emph{explicitly} store and recall past experiences for use in future decision-making. 
The concept, ``working memory'', originates from cognitive psychology and neuroscience \citep{baddeley2003working, goldman1995cellular},
where it refers to the system responsible for the temporary storage and manipulation of information during cognitive tasks.

Our motivation comes from how humans think before they act: they can reason on past experiences to generate appropriate behavior in new situations.
We want to equip our robots with similar abilities.
Imagine training a robot to play four different Atari games: Asteroids, Asteroids Deluxe, Space Invaders, and Space Invaders II (Figure~\ref{figs:intro}).
Asteroids Deluxe is a sequel to Asteroids that introduces new boss fights and enemies, similarly, Space Invaders II is a sequel to Space Invaders.
For a robot to play these four games, it must actively store what it has learned in memory and choose the appropriate strategy for each game.
Throughout training, the robot's memory module continuously processes and updates relevant game information, allowing it to make informed decisions and adapt its strategies.

Followed by this intuition, we introduce \fullname~
(\name): 
which represents a working memory as a matrix, the functioning of the memory entails two primary steps, namely \textbf{memory update} and \textbf{memory retrieval}. 
\name builds on earlier work on memory-augmented neural networks~\citep{santoro2016meta}---including neural Turing machines~\citep{graves2014neural} and memory networks~\citep{sukhbaatar2015end}---in several ways, as we detail in Section~\ref{section:related_work}. 

We use content-based addressing \citep{DBLP:conf/nips/EslamiHWTSKH16} to locate the memory position to update information into or retrieve information from it.  
The memory update involves modifying or replacing existing information. 
This enables the system to keep track of changes, maintain task-relevant information, and facilitate decision-making.
More specifically, we first map both the input sequence and memory into three entities: \emph{query}, \emph{key}, and \emph{value}. 
Next, we use an attention-based mechanism to calculate the correlations between the input and memory, and subsequently use the attended weight of the input sequence to update the memory.
Memory retrieval refers to the process of accessing and recovering stored information. It involves bringing relevant information back to condition decision-making.
We read from the updated memory at the content-based address to achieve that.

Humans can make analogies by mapping experience between tasks, this enables us to leverage experience when encountering new tasks.
Therefore, we also equip our memory module with an adaptable mapping capability.
Specifically, for adapting the memory module to a new task, we employ the Low-Rank Adaptation (LoRA) method as described in \citep{DBLP:conf/iclr/HuSWALWWC22} as our fine-tuning strategy.
The main idea behind LoRA is to train a low-rank projection matrix on a small amount of labeled data from a new task. 
This matrix maps the parameters of a pre-trained model to a new task.
In this work, we fine-tune only the memory module because we rely on the generalization capacity of a pre-trained Decision Transformer (DT). 
In prior works, Transformers are often pre-trained on large-scale datasets \citep{DBLP:conf/nips/LeeNYLFGFXJMM22,DBLP:journals/corr/abs-2304-08487},
such pre-training enables them to capture broad knowledge that is transferable across tasks. 
In contrast, the memory module we propose is designed to store knowledge explicitly that can be modified and utilized to new tasks.

\name differs from external memory and information retrieval-based methods in several ways. 
First, external memory methods often require a large dataset that serves as a look-up table; whereas in our system, the working memory can be maintained at a handy size. 
Second, external memory methods require an extra step of representation learning to convert entries in the look-up table into vector space; in contrast, information in our system naturally operates in a latent space.
Finally, external/retrieval-based memory methods often resort to a $k$-nearest neighbor search during retrieval, mainly because it is computationally impractical for an attention-based mechanism to operate on large sets (the look-up table); our method, on the other side, could naturally leverage attention-based module during both memory update and retrieval.

To validate our approach, we evaluate \name in two environments and compare against a set of strong baselines: 
(a) Atari games: we compare against Multi-game Decision Transformer \citep[MDT,][]{DBLP:conf/nips/LeeNYLFGFXJMM22} and Recurrent Memory Decision Transformer \citep[RMDT,][]{bessonov2023recurrent};
(b) Meta-World environments: we compare against Prompt Decision Transformer \citep[PDT,][]{DBLP:conf/icml/XuSZLZTG22} and Hyper-Decision Transformer \citep[HDT,][]{DBLP:journals/corr/abs-2304-08487}.
Our results show that \name improves generalization and adaptability with fewer model parameters and less training time.

\section{Related work}
\label{section:related_work}

\textbf{Transformer-based Reinforcement Learning methods}
Transformer \citep{DBLP:conf/nips/VaswaniSPUJGKP17} is a powerful architecture designed for sequence modeling. 
Owing to the capabilities that emerge as model and data size scale up, the Transformer has become a foundational model in several domains, including natural language processing  \citep{DBLP:journals/corr/abs-2005-14165, DBLP:journals/corr/abs-2303-08774, DBLP:journals/corr/abs-2302-13971} and computer vision \citep{DBLP:conf/iclr/DosovitskiyB0WZ21}. 
However, applying Transformer in reinforcement learning settings, such that it generalizes to multiple tasks, remains an open problem.

Recently, \citet{DBLP:conf/nips/ChenLRLGLASM21} and \citet{janner2021sequence} treat the RL problem as a sequence modeling problem and proposed a Transformer-based architecture to solve it with offline RL. 
These findings inspired researchers to develop more advanced Transformer-based RL methods. 
Subsequent efforts mainly focus on two aspects: generalization and adaptability.
To improve model online adaptability, \citet{DBLP:conf/icml/ZhengZG22} propose the Online Decision Transformer (Online DT), which utilizes the maximum-entropy idea to encourage pre-trained policies to explore during a phase of online adaptation. 
To improve offline adaptation, \citet{DBLP:journals/corr/abs-2304-08487} propose a Hyper-network-based module that helps DT adapt to unseen tasks efficiently.
To facilitate task adaptation, \citet{DBLP:conf/icml/XuSZLZTG22} introduce the prompt-based DT, which selects short trajectories to use in a task prompt in analogy with in-context learning for large language models.
Furthermore, \citet{DBLP:conf/nips/LeeNYLFGFXJMM22} propose a multi-game DT (MDT), which use the expert action inference to produce actions of highly-rewarding behavior consistently. MDT demonstrates that DT can generalize to various Atari games with human-level performance.

We argue that the generalization of the above-mentioned works relies on the size of the models and does not learn the data efficiently.
To address this issue, we introduce a memory module that can store, mix, and retrieve training information for better model and training efficiency.

\textbf{Working memory}
In the context of machine learning, there is a long history of neural network-based models that incorporate memory mechanisms~\citep{das1992learning,schmidhuber1992learning,hochreiter1997long,santoro2016meta,ba2016using,munkhdalai2017meta,pmlr-v80-munkhdalai18a,csordas2019improving,ramsauer2020hopfield,wu2022memvit}.
Generally, this research aims to enhance the capacity of neural networks to store and manipulate information over extended periods of time, leading to improved performance on a range of tasks. It often takes inspiration from human cognitive function.
Most salient to our work, \citet{graves2014neural} merge concepts from Turing machines and deep learning in ``Neural Turing Machines'' (NTMs), neural networks that include a content-addressable matrix memory space for storing and updating information throughout time. 
They show NTMs to be effective for various algorithmic tasks.
Concurrently, \citet{sukhbaatar2015end} introduce ``memory networks,'' which use a content-addressable matrix memory to store and retrieve information from previous computational steps, facilitating complex reasoning and inference tasks. 

Infinity-former excels in handling unbounded contexts with precision and flexibility, ideal for extensive and complex datasets \citep{DBLP:conf/acl/MartinsMM22}.
LONGMEM decoupled architecture and token-to-chunk retrieval make it adept at managing large contexts and overcoming memory staleness \cite{wang2024augmenting}.
kNN-augmented Transformer offers flexibility in context length and rapid adaptation to new data, enhancing the model's real-time applicability \cite{wu2022memorizing}.

More recently, \citet{bessonov2023recurrent} introduces a recurrent memory mechanism to address reinforcement learning challenges, which preserves a hidden state throughout the decision-making process. However, this method overlooks the storage and retrieval of task-related information, thereby falling short of fostering model generalization and task adaptation.
\citet{munkhdalai2019metalearned} propose a rapidly adaptable neural memory system, with which they instantiate as a feedforward neural network trained by meta-learning \cite{DBLP:CanBi}. They evaluate the memory's effectiveness in a simple RL setting, maze exploration, and on various NLP tasks.
Alternatively, \citet{DBLP:conf/iclr/GoyalDLBKRBBMB22} builds on the ``global workspace'' theory from cognitive science, which posits that different input entities share information through a common communication channel. 
The proposed shared global workspace method employs the attention mechanism to encourage the most useful information to be shared among neural modules. 
It is closely related to working memory and inspires us to explore how an explicit working memory can improve the generalization of Transformer-based models.
An upshot of our work is that it may be valuable to revisit earlier memory-augmentation methods in light of more powerful foundation models.

\textbf{Comparisons with classic memory-based methods}
There are several classic memory-based methods; however, none of them can be directly applied to DT in the same manner as RMDT. Consequently, we have not included them in our evaluations. To distinguish our approach, we compare these methods in terms of both memory writing and memory reading.
\textbf{Memory Writing:}
$\infty$-former \cite{DBLP:conf/acl/MartinsMM22} represents memory as a continuous signal using radial basis functions (RBFs). When new information is encountered, it is integrated into this continuous representation. This process involves evaluating the continuous signal at specific locations and then concatenating these evaluations with new vectors coming from the short-term memory.
DT-Mem involves sophisticated mechanisms using attention to manage the significance of new and existing data. This process would be based on calculating correlations between the input and memory and updating the memory with the attended weight of the input sequence.
LONGMEM \cite{wang2024augmenting} caches paired attention keys and values from the previous context into a non-differentiable memory bank using a frozen backbone Large Language Model (LLM) as the memory encoder.
KNN-transformer \cite{fan2021augmenting} uses (key, value) pairs from the local context are appended to the end of an external memory.
\textbf{Memory Reading:}
The reading process of $\infty$-former utilizes a continuous-space attention framework.
DT-Mem uses content-based addressing for memory retrieval. This process would involve using attention mechanisms to read from the updated memory, focusing on the content relevant to the current task or context.
LONGMEM uses a decoupled memory module, specifically a residual side-network (SideNet), for memory retrieval and reading. The SideNet retrieves cached key-value pairs of previous contexts from memory using the attention query of the current input.
KNN-transformer features a kNN-augmented attention layer that combines standard dense self-attention with approximate k-nearest-neighbor search into the external memory. The kNN lookup retrieves the top-k (key, value) pairs for each query from the input subsequence, constructing an attention matrix that represents these memories differently for each query.

\section{Preliminaries}

\subsection{Offline Reinforcement Learning}
\label{section:offline_rl}


A trajectory consists of a series of states, actions, and rewards, expressed as $\tau = (s_0, a_0, r_0, s_1, a_1, r_1, \cdots, s_T, a_T, r_T)$. In the context of offline RL, data acquisition does not come from active interaction with the environment. Instead, we rely solely on a predefined and limited dataset containing various trajectories generated by different policies. This scenario presents greater challenges as it restricts the agent's ability to actively explore the environment and gather new information, which is a crucial aspect of traditional RL approaches.

Formally, in the context of model evaluation, we can define a set of training tasks and testing tasks as $T^{\text{train}}$ and $T^{\text{test}}$, respectively. 
These two sets deliberately have no overlapping tasks but may share the same or similar observation and action spaces.
To be more specific, for each training task $\mathcal{T}^i \in T^{\text{train}}$, we have access to a large training dataset, which contains trajectories $\tau^{0:H} = (s_0, a_0, r_0, \cdots, s_H, a_H, r_H)$, where $H$ is the episode length. 
However, we assume access to only a small amount of data for the testing tasks.

Our goal is to evaluate the proposed model in two dimensions. 
First, we want to assess the model's \textbf{generalization}, which refers to its ability to solve the testing tasks within a finite time with no additional fine-tuning. 
Second, we want to test the model's \textbf{adaptability}, which refers to its ability to improve its performance on the testing tasks through fine-tuning on limited data after pre-training on separate tasks.

\subsection{Low-rank Adaptation}

Low-rank adaptation \citep[LoRA,][]{DBLP:conf/iclr/HuSWALWWC22} is a transfer learning technique used to adapt a pre-trained model to a new task with limited labeled data. 
The main idea behind LoRA is to utilize a small amount of labeled data from a new task to learn a low-rank projection matrix. This matrix maps the parameters of a pre-trained model to the new task.
Specifically, for a pre-trained weight matrix $\bm{W}\in \mathbb{R}^{d\times k}$, we assume a low-rank decomposition for the weight update: $\bm{W}+\Delta\bm{W}=\bm{W}+\bm{B}\bm{A}$, where $\bm{B}\in  \mathbb{R}^{d\times r}$ and $\bm{A}\in  \mathbb{R}^{r\times k}$.
Once the projection matrix is learned, it can transform the pre-trained model's parameters to a new subspace that is more suitable and which alters its forward pass output. 
In other words, $\bm{W}\bm{x}+\Delta\bm{W}\bm{x}=\bm{W}\bm{x}+\bm{B}\bm{A}\bm{x}$.

\begin{figure*}[t!]
    \centering
    \includegraphics[width=0.7\textwidth]{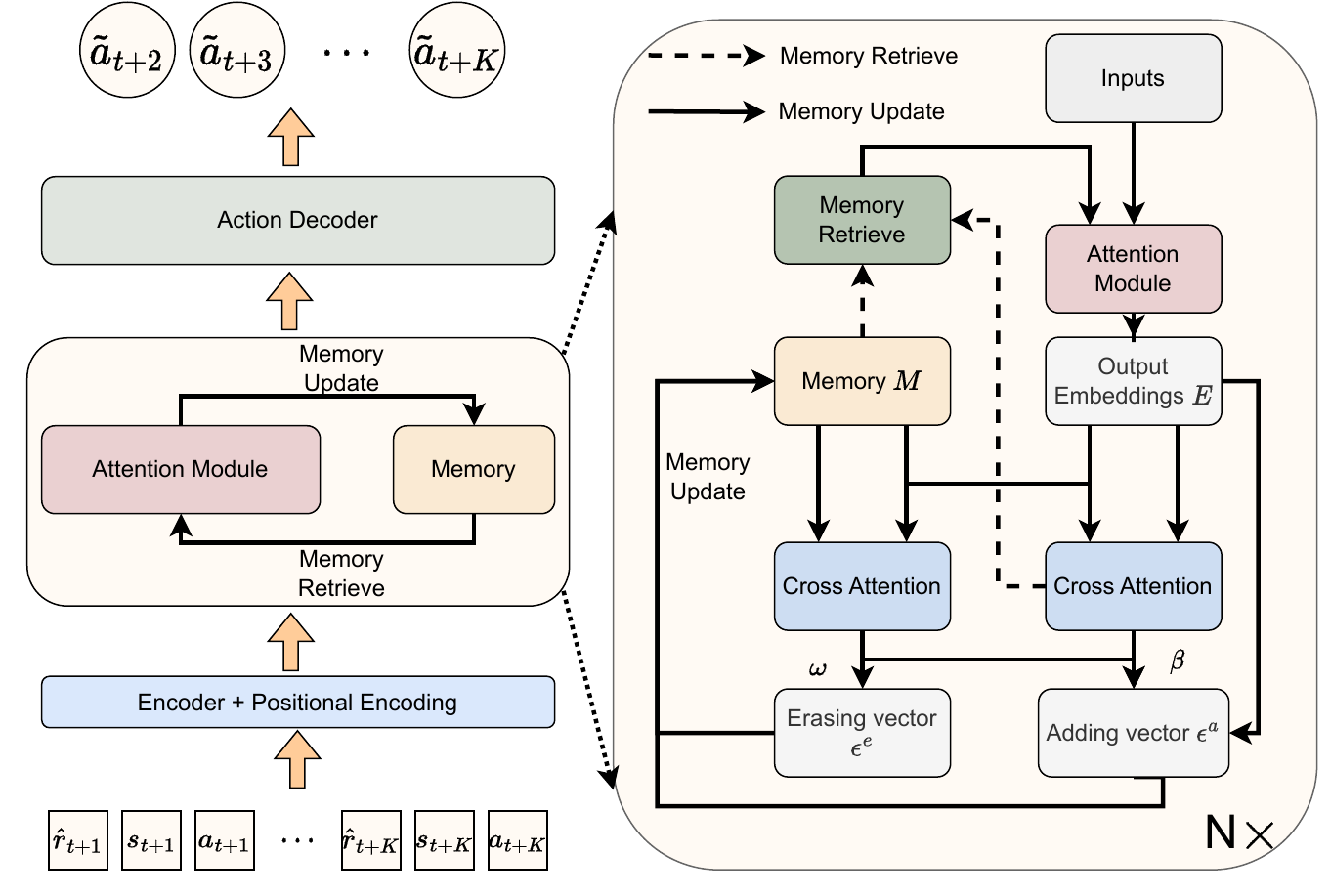}
    \caption{An overview of the proposed \name architecture. The input of the encoder is a fixed-length sequence of trajectories. The encoder with positional encoder module embeds the inputs and persists the temporal correlations between states and actions. The primary role of the attention module is to capture dependencies and relationships between states, actions, and returns in a sequence. Note that there are multiple attention modules stack together. Our design deconstructs this module and manages the memory flows between the attention module within each block. The output from attention blocks flows to the action decoder, which decodes back to the real actions.}
    \label{fig:archi}
\end{figure*}

\section{Methodology}

\subsection{Overview}


In Figure~\ref{fig:archi}, we depict the architecture of \name, which consists of four components: the attention module, the encoder module associated with positional encoding, the action decoder module, and the memory module.
The primary role of the attention module is to capture dependencies and relationships between states, actions, and returns in a sequence.
The encoder with positional encoder module embeds the inputs and persists the temporal correlations between states and actions.
The input of the encoder module is a fixed-length sequence of trajectories, denoted as $\tau^{t+1:t+K}$. The output is a sequence of embeddings, where each entry can be attended state embeddings, action embeddings, or return-to-go embeddings. 
The action decoder is a multi-layer perceptron (MLP) that responds to decode the latent parameters of actions.
We introduce a memory module for storing and manipulating intermediate information on top of the standard Decision Transformer architecture. 
This memory design is inspired by the Neural Turing Machine \citep{graves2014neural}, where the memory is utilized to infer multiple algorithms.
The details of memory module design and functionality are introduced in Section~\ref{subsec:mem}.

\subsection{Memory Module}
\label{subsec:mem}

The design for the memory module is inspired by the way humans think before they act. 
Its functioning consists of three parts: identifying salient information output from the Transformer module; determining where to store new information and how to integrate it with existing memory slots; and considering how to use these memory slots for future decision-making. 
We break down these questions and design the following steps to address them.

\textit{\textbf{Step 0: Memory Module Initialization.}} 
The memory is initialized as a random matrix $\bm{M}$, where each row $\bm{m_i}\in \mathbb{R}^d$, with $i\in[0,N]$, represents a memory slot. 

\textit{\textbf{Step 1: Input Sequence Organizing.}} 
As illustrated in Section~\ref{section:offline_rl},an input sequence comprises multiple steps of the tuple \(<\hat{r}_t, s_t, a_t>\). Instead of directly feeding this sequence into the Transformer module, we treat each tuple as an entity and embed them into the same latent space. Specifically, we define embedding functions \(g_s(s) = \bm{e_s}\), \(g_a(a)=\bm{e_a}\), and \(g_r(\hat{r})=\bm{e_{\hat{r}}}\), where \(\bm{e}_s\), \(\bm{e}_a\), and \(\bm{e}_{\hat{r}} \in \mathbb{R}^d\) with \(d\) representing the dimension in the latent space. The final input sequence emerges from the concatenation of embeddings \(\bm{E}=[\cdots, [\bm{e}_{s_t};\bm{e}_{a_t};\bm{e}_{\hat{r}_t}],\cdots]\).

Recall that we represent our memory as a matrix with fixed dimensions (i.e., number of slots $\times$ hidden dimensions). It is crucial to synchronize the input dimensions for efficient storage. Notably, in our design, we maintain the relationships among trajectories as posited in the Decision Transformer paper, although this is not a requisite. For instance, in another prior work, Trajectory Transformer \cite{janner2021sequence}, states, rewards, and others are grouped individually. 
We demonstrate in Appendix~\ref{subsec:input_organize} that these varied designs exhibit no significant difference.

\textit{\textbf{Step 2: Content-based Address.}} 
We use an attention-based mechanism to locate a proper memory slot for new input by identifying correlated information. 
This approach is based on the idea that humans tend to store and group similar or related information together (e.g., in Documentation Science and Archival Science \citep{DooleyArchivalAdvantage}). 
To locate the memory position, we compute the position address $\bm{w}$ as follows:
$\bm{w} = \text{softmax}\Big(\frac{\bm{Q}\bm{K}^T}{\sqrt{d}}\Big)$. Here, $\bm{Q}=\bm{M}\bm{W}^q$ and $\bm{K}=\bm{E}\bm{W}^k$, where $\bm{W}^q$ and $\bm{W}^k$ are parameters in the MLP. 
The objective is to map the memory and input information into the query and key matrix, and then use the dot product to determine the similarities between these two matrices. 
The softmax function guarantees that the sum of all addresses equals one.

\textit{\textbf{Step 3: Memory update.}}
To store incoming information and blend it with existing memory, 
we calculate two vectors: an erasing vector, $\bm{\epsilon}^e$, and an adding vector, $\bm{\epsilon}^a$.
The erasing vector erases the current memory, while the adding vector controls information flow to the memory.
To achieve this goal, we again utilize the attention mechanism.
First, we map memory and input information to query, key, and value vectors, denoted as $\bm{\hat{Q}}=\bm{M}\bm{\hat{W}}^q$, $\bm{\hat{K}}=\bm{E}\bm{\hat{W}}^k$, and $\bm{\hat{V}}=\bm{E}\bm{\hat{W}}^v$, respectively, where $\bm{\hat{W}}^q$, $\bm{\hat{W}}^k$, and $\bm{\hat{W}}^v$ are parameters.
Next, we calculate the writing strength, $\beta = \text{softmax}\Big(\frac{\bm{\hat{Q}}\bm{\hat{K}}^T}{\sqrt{d}}\Big)$.
The erasing vector is used to selectively erase information from the memory matrix and is computed as a function of the content-based addressing vector and the write strength.
The erasing vector is calculated as $\bm{\epsilon}^e = \bm{w}\odot(1-\beta)$, where $\odot$ indicates element-wise multiplication.
The complement of the write strength is 1 minus the write strength, so this will result in a vector where the elements corresponding to the selected memory locations are set to 0, and the elements corresponding to the unselected memory locations are unchanged.

The adding vector is used to selectively add information to the memory matrix and is computed as a function of the write strength and the input vector.
Specifically, the adding vector is calculated as $\bm{\epsilon}^a=(\bm{w}\odot\beta) \bm{\hat{W}}^vx$.

Finally, the memory is updated as $\bm{M}_t = \bm{M}_{t-1}\odot(\bm{1}-\bm{\epsilon}^e)+\bm{\epsilon}^a$.
The new information will be stored if the selected memory slot is empty or erased.
Otherwise, the new information will be blended with the existing memory contents.

\textit{\textbf{Step 4: Memory retrieval.}}
We retrieve information from the updated memory slots to utilize memory for decision-making. 
Reading from the memory matrix is done by computing a read position vector. 
This vector can be computed using the above content-based addressing mechanism that compares the query vector with the contents of the memory matrix. 
Note that in other retrieval-based methods \citep{DBLP:conf/nips/HumphreysGTSWL22,DBLP:conf/icml/BorgeaudMHCRM0L22}, the nearest neighbor is the common way to retrieve related information. 
However, in our case, the working memory is considerably smaller than typical external memory, which makes attention-based retrieval feasible. 
Since the query information is the same as the input information, we use the same content address to retrieve the memory: $\bm{E}_{\text{out}} = \bm{w}\odot\bm{M}_t$.

\subsection{Pre-training}
\label{subsec:pretrain}

We use a set of training tasks $T^{\text{train}}$, where each task $\mathcal{T}_i\in T^{\text{train}}$ has an associated offline dataset $\mathcal{D}_i$ consisting of hundreds of trajectories $\tau$ generated by a behavior policy. 
The behavior policy can either be a pre-trained policy (such as DQN), a rule-based policy, or even human players --- depending on what is available. 
Each trajectory $\tau = (s_0,a_0,r_0,\cdots,s_H,a_H,r_H)$, where $s_i\in \mathcal{S}, a_i\in \mathcal{A}, r_i\in \mathcal{R}$, and $H$ is the episode length.

To serve as an input to the \name, we first segment the trajectory $\tau$ into several pieces, each with length $K$.
We denote $\tau_{t+1:t+K} = (s_{t+1},a_{t+1},r_{t+1},\cdots,s_{t+K},a_{t+K},r_{t+K})$ as one of the input sequence.
However, we modify these trajectories instead of inputting them directly.
Specifically, we follow the return-to-go Decision Transformer idea \cite{DBLP:conf/nips/ChenLRLGLASM21} and calculate the return to go, $\hat{r}_t = \sum^{t+K}_{t+1}r_t$, which is performed at every timestep.
This is effective because $\hat{r}_t$ acts as a subgoal, encouraging the Transformer module to generate actions that can reduce the negative of this value as close to zero as possible.
Therefore, we feed the modified trajectories $\hat{\tau}_{t+1:t+K} = (\hat{r}_{t+1},s_{t+1},a_{t+1},\cdots,\hat{r}_{t+K},s_{t+K},a_{t+K})$ as input to the Transformer module.
The output of the Transformer module is a sequence embedding $\bm{e}_{\text{seq}}\in \mathbb{R}^{d\times 3K}$, where $d$ is the dimension of the embedding space.

Next, we transmit $\bm{e}_{\text{seq}}$ to the memory module to update and retrieve the memory information. 
Finally, we feed the retrieved information $\bm{E}_{\text{out}}$ into the MLPs to generate the corresponding actions $\hat{a}_t$. 
We minimize a supervised training loss with three terms: predicted actions $\Tilde{a}_t$, predicted reward $\Tilde{r}_t$, and predicted return-to-go $\Tilde{R}_t$. The loss function is:
\begin{equation}
\mathcal{L} = \sum_{t+1}^{t+K}||\Tilde{a}_t - a_t||^2+\alpha ||\Tilde{r}_t-\hat{r}_t||^2 + \lambda||\Tilde{R}_{t}-r_t||^2,
\label{eqn:loss}
\end{equation}
where $\alpha$ and $\lambda$ are scalar hyper-parameters. 
Empirically, we find that the final performance is not sensitive to specific $\alpha$ and $\lambda$ values, so we set them to 1 for simplicity.

The full pre-training process is summarized in Appendix \ref{subsec:train_algo} Algorithm~\ref{alg:alg1}.

\begin{table*}[htbp]
\centering
\begin{tabular}{c|c|c|c|c|c|c}
\hline
Model    & Layers & Hidden size (d) & \multicolumn{1}{c|}{Heads} & Params & Memory Size               & Memory Module Params \\ \hline
HDT      & 4      & 512             & 8                          & 13M    & N.A.                      & N.A.                 \\ \hline
MDT-200M & 10     & 1280            & 20                         & 200M   & N.A.                      & N.A.                 \\ \hline
\name     & 4      & 512            & 8                         & 13M    & 559K                      & 7M                   \\ \hline
\end{tabular}
\caption{Implementation details of model sizes}
\label{tab:model_sizes}
\end{table*}

\begin{table*}[t!]
\centering

\begin{tabular}{c|cccc}
\hline
          & Alien                                                             & MsPacman                                                                                                               & SpaceInvaders                                                    & StarGunner                                                       \\ 
MDT       & 3.8\% ($\pm 0.4\%)$     & 13.2\% $(\pm 1.3\%)$   & 8.6\% $(\pm 1.6\%)$    & 2.3\% $(\pm 0.1\%)$    \\ 
RMDT      & 22.3\% $(\pm 10.7\%)$   & 22.9\% $(\pm 8.9\%)$ & 17.6\% $(\pm 9.2\%)$   & 27.7\% $(\pm 11.5\%)$ \\ 
DT-Mem    & 51.0\% $(\pm 32.2\%)$  & 69.3\% $(\pm 19.3\%)$  & 53.6\% $(\pm 29.0\%)$  & 62.2\% $(\pm 19.1\%)$  \\ \hline
\end{tabular}

\caption{Evaluation results on 4 held-out games after pre-training on other Atari Games. Each value represents the DQN-normalized score, computed with a 95\% confidence interval.
}
\label{tab:dqn_score}
\end{table*}

\subsection{Fine-tuning with LoRA}
\label{subsec:fine-tune-lora}

Fine-tuning LLMs involves heavy computation due to the large number of parameter updates required. 
We argue that fine-tuning only the memory module can achieve results comparable to those of fine-tuning the entire parameter space. 
LLMs such as BERT \citep{DBLP:conf/naacl/DevlinCLT19} or GPT \citep{radford2019language} greatly benefit from training on large-scale datasets, which expose the model to a diverse range of linguistic patterns and semantic relationships. 
This exposure helps the model learn robust and generalized representations that capture different language understanding and generation aspects. 
After pre-training, the model can be fine-tuned on specific downstream tasks with task-specific labeled data. 
In our case, this task-specific knowledge is stored in the memory module. 
Thus, fine-tuning the memory module helps the model update its memory module to adapt to the new task.

We apply the low-rank adaptation approach \citep[LoRA,][]{DBLP:conf/iclr/HuSWALWWC22} to fine-tune the memory module. 
Specifically, we modify the forward pass by adding low-rank matrices to $\bm{W}^q$, $\bm{W}^k$, $\bm{W}^v$, $\bm{\hat{W}}^q$, and $\bm{\hat{W}}^k$.
Taking $\bm{W}^q$ as an example. 
Assuming the original output for query information is $\bm{Q}=\bm{M}\bm{W}^q$, we adapt this query value to a new task as $\bm{Q'}=\bm{M}(\bm{W}^q+\bm{B}^q\bm{A}^q)$, where $\bm{W}^q\in\mathbb{R}^{n\times d}$, $\bm{B}\in\mathbb{R}^{n\times m}$, and $\bm{A}\in\mathbb{R}^{m\times d}$, and $m$ is the size of the memory module.
Since the rank $m\ll \text{min}(n,d)$, fine-tuning the parameters $\bm{B}^q$ and $\bm{A}^q$ reduces the number of trainable parameters for downstream tasks.
We perform supervised training by computing the loss between the model's output and the labels in the fine-tuning dataset. 
Only $\bm{B}^q$ and $\bm{A}^q$ are updated during this process.
The detailed fine-tuning procedure can be found in Appendix~\ref{subsec:train_algo} Algorithm~\ref{alg:alg2}.

\section{Evaluation}

We design our experiments to answer the following questions:
\begin{itemize}
    \item \textbf{Q1:} Does \name improve model generalization?
    \item \textbf{Q2:} Does \name improve pre-training results and training efficiency?
    \item \textbf{Q3:} Does \name scales with model size?
    \item \textbf{Q4:} Does fine-tuning only the memory module improve model adaptability?
\end{itemize}

We use generalization to refer to performance on tasks the model has never been trained on (zero-shot), and adaptability to refer to performance after fine-tuning.

\subsection{Environments and Models Setup}

\textbf{Atari Games.} To ensure a fair comparison with the Multi-Game Decision Transformer, we use the same Atari dataset,
which comprises multiple training runs of DQN trajectories. 
Due to limited compute resources and to prevent cherry-picking, we select 17 games from a total of 41 based on their alphabetical order, as introduced in \cite{DBLP:conf/nips/LeeNYLFGFXJMM22}.
For each game, the data contains 50 policy checkpoints, each containing 500k environment steps.
For the fine-tuning dataset, we randomly selected 10\% of the data from the unseen dataset, yielding 50k environment steps.
Following the settings from \citet{DBLP:conf/nips/LeeNYLFGFXJMM22}, we choose four games (Alien, Ms. Pac-Man, Space Invaders, and Star Gunner) to be used only for fine-tuning.
Moreover, \citet{DBLP:conf/nips/BrandfonbrenerB22} suggests that return-conditioned supervised learning (RCSL) algorithms require strong dataset coverage to select a near-optimal policy. 
Therefore, our dataset contains both expert and non-expert behaviors.

\textbf{Meta-World.} To make a fair comparison with Hyper-DT and Prompt-DT, we evaluate the proposed method on the Meta-World environment \citep{yu2019meta}. 
We evaluate using the Meta-World ML45 benchmark, which includes 45 training tasks and 5 testing tasks.
Following the approach taken in \cite{DBLP:journals/corr/abs-2304-08487}, for each training task, we generate an offline dataset containing 1000 episodes for each game, using a rule-based script policy.
For fine-tuning data, we randomly pick 10k episodes from the testing dataset, as compared to 20k-80k episodes used in Hyper-DT.

\textbf{Implementation details.} We report results for \name 20M (20 million parameters), which consists of 13M Transformer parameters and 7M memory module parameters. 
For all games, we use eight V100 GPUs for model training and one V100 GPU for fine-tuning.
We train on both Atari games and Meta-World for 10M steps. For fine-tuning on unseen scenarios, we train for 100k steps.

Table \ref{tab:model_sizes} summarizes the different model configurations used for evaluation.
In this section, we describe these model configurations in detail. While Table \ref{tab:model_sizes} provides a summary, we will provide additional information here.
\name, PDT, and HDT all share the same Transformer architectures. However, for task-adaptation, HDT utilizes a pre-trained 2.3M hyper-network, while \name introduces 147K LoRA parameters.
To compare with MDT, we use the same parameter size as reported in \cite{DBLP:conf/nips/LeeNYLFGFXJMM22}.

\subsection{Baseline Methods}

We compare \name's performance against the following baselines.
\textbf{MDT}: Multi-game Decision Transformer \citep{DBLP:conf/nips/LeeNYLFGFXJMM22}, which trains a large Transformer-based model on multi-game domains.
For a fair comparison, we train an MDT with 20M parameters, which is approximately the same size of DT-Mem.
\textbf{RMDT}: Recurrent Memory Decision Transformer \citep{bessonov2023recurrent}, which utilizes a recurrent memory mechanism for solving reinforcement learning problems. This is the most related memory-based DT that is close to our work.
\textbf{HDT}: Hyper-Decision Transformer \citep{DBLP:journals/corr/abs-2304-08487}, which utilizes a hyper-network module to help DT adapt rapidly to unseen tasks.
Since we do not have access to the implementation at the time of writing, for the sake of correctness, we compare our model with HDT on Meta-World only.
The results reported in our evaluation section come from the HDT paper.
\textbf{PDT}: The Prompt Decision Transformer \citep{DBLP:conf/icml/XuSZLZTG22} generates actions by considering both recent context and pre-collected demonstrations from the target task. 
\color{black}

\subsection{\name improves model generalization.}

We evaluate four held-out games fine-tuning results as listed in Table~\ref{tab:dqn_score}. 
Each evaluation signifies an average derived from 16 runs, each under differing random seeds. 
The derived results show that the memory-incorporated methods, RMDT and \name, enhance model generalization when compared to their ablation method MDT. 
A noteworthy observation is that \name demonstrates superior generalization performance than RMDT in all four games. 
This suggests that RMDT does not adequately address the storage and retrieval of information related to tasks, consequently not effectively supporting model generalization and task adaptation development. This contrasts DT-Mem, which is designed to accomplish these objectives.


\subsection{\name enables more computationally efficient training and scale with model parameters.}

To demonstrate training efficiency, we illustrate the model training time in Table~\ref{tab:train_time} and the training curve in Appendix \ref{subsec:train_eff} Figure \ref{fig:train_eff}.
During training, we find that \name reduces the training time by approximately 4 times, 8 times, and 32 times compared to MDT-13M, MDT-40M, and MDT-200M, respectively.
It is reasonable to report the prediction loss on the training dataset for the training curve since we use a supervised loss.
As defined in Section~\ref{subsec:pretrain}, the prediction accuracy consists of three parts: action prediction accuracy, reward prediction accuracy, and return prediction accuracy.

\begin{figure}[t!]
    \centering
        \centering
        \includegraphics[width=.5\textwidth]{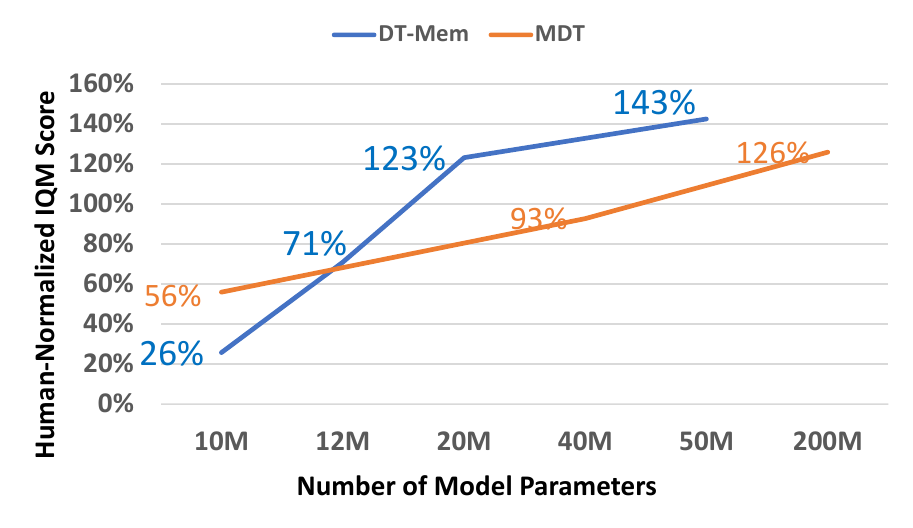} 
        \caption{Scaling of IQM scores}
        \label{fig:scale_law}
\end{figure}

\begin{table}
    \centering
    \begin{tabular}{c|c}
        \hline
        Model           & Training time (hours) \\
        \textbf{DT-Mem-20M} & \textbf{50}           \\
        MDT-13M         & 200                   \\
        MDT-40M         & 400                   \\
        MDT-200M        & 1600                  \\ \hline
        \end{tabular}
    \caption{Model training time}
    \label{tab:train_time}
\end{table}

Figure ~\ref{fig:scale_law} showcases the scaling laws of the proposed \name model. 
We measure performance using the human-normalized IQM score. It's crucial to note that for all instances of \name, we maintained a consistent number of memory slots.
The result shows that the performance of \name scales with the number of parameters. 
Notably, the generalization of \name with 20M parameters is approximately on par with MDT's 200M parameter version. 
Furthermore, the 50M \name surpasses MDT by a margin of 16.7\%.

\subsection{Fine-tuning only the memory module improves model adaptability.}
\label{subsec:fine-tune}

\begin{figure*}[ht]
    \centering
    \includegraphics[width=0.8\textwidth]{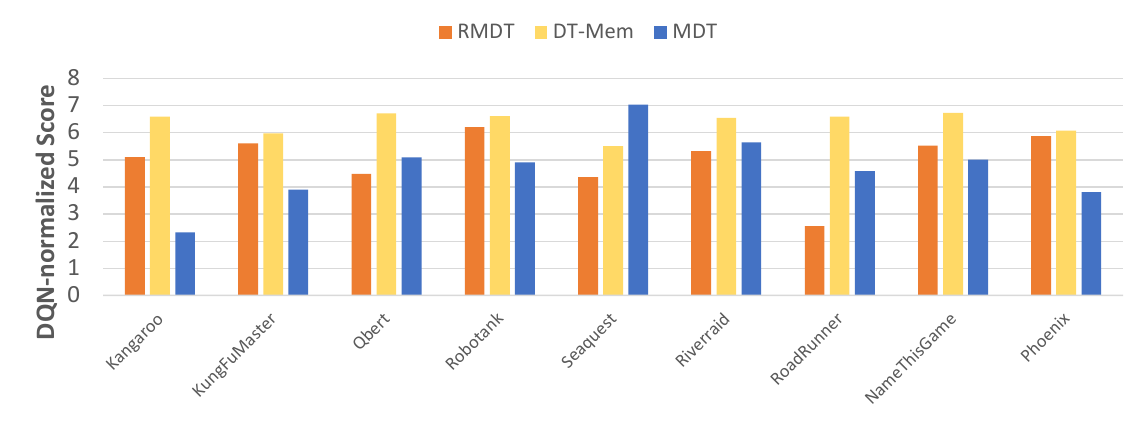}
    \includegraphics[width=0.8\linewidth]{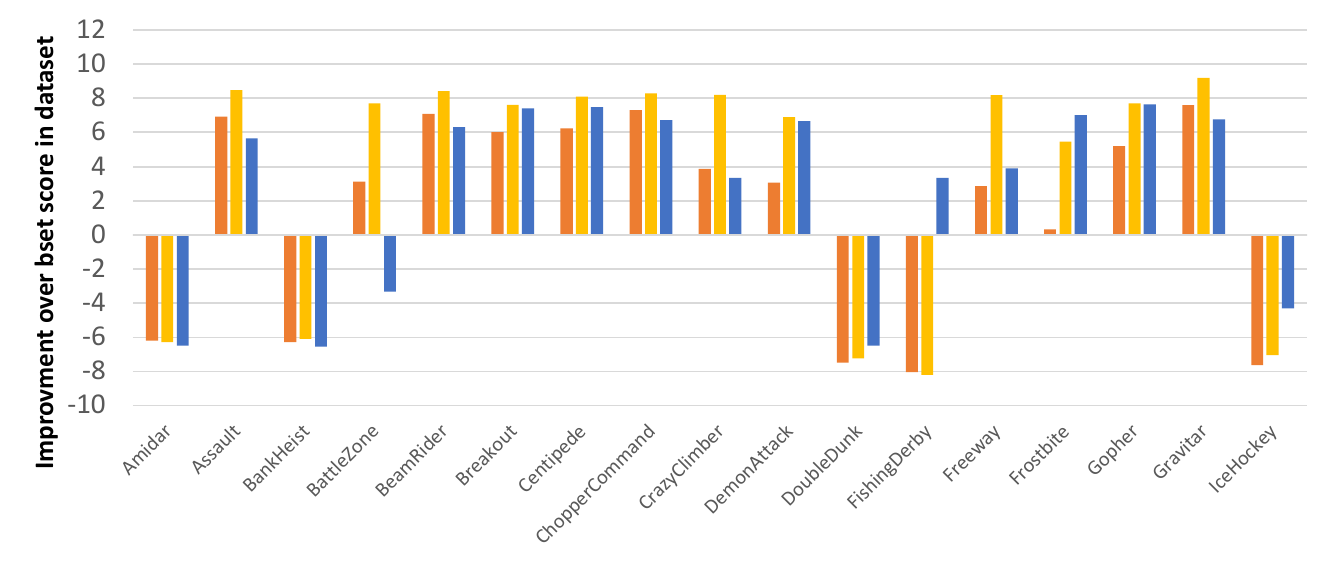}
    \caption{\textbf{Top}: Fine-tuning performance on 10\% of dataset in unseen Atari games. For better visualization, the y-axis is the logarithm of DQN-normalized score. \textbf{Bottom}: The performance improvement for the training dataset.}
    \label{fig:adapt}
\end{figure*}

Another question we care about is how the pre-trained \name performs on unseen tasks. 
We randomly selected nine unseen Atari games and evaluated their performance through relative improvement scores, as shown in Figure~\ref{fig:adapt} (top). 
\name consistently outperforms RMDT and MDT in most of the games listed, with the exception of Seaquest, where MDT excels.
MDT exhibits the least superior performance across most games, with its performance particularly lagging in KungFuMaster, Robotank, and Phoenix.
RMDT holds an intermediate performance level between \name and MDT across most games.
The consistent superior performance of \name across most games suggests that this method might have a more adaptable approach.
The singular superior performance of MDT in Seaquest prompts a further investigation into the unique attributes of this game that may favor the MDT method.

To further understand the adaptability of the proposed method, we compare \name with HDT and PDT in Meta-World environments.
The quantitative fine-tuning results are shown in Table~\ref{tab:meta45}.
Overall, \name achieves the best performance in the comparison.
As we can see, compared to HDT, \name increases training, testing (no-FT), and testing (FT) scores by an average of 3\%, 8\%, and 3\%, respectively.
Moreover, the HDT adaptation module (hyper-network module), while small (69K) relative to the full model (13M), relies on the pre-trained hyper-network, which contains 2.3M parameters. We argue that the hyper-net is more burdensome than our design: it uses more than 10x the number of adaptation parameters (147K) used by \name and requires an extra compute phase to pre-train the hyper-network module, which again could cause extra issues in optimization.

\begin{table*}[htbp]
\small
\centering
\begin{tabular}{c|cc|ccc}
\hline
      & \multicolumn{2}{c|}{Model Sizes} & \multicolumn{3}{c}{Meta-World ML45 Performances}                                                                   \\
      & Adaptation      & Percentage     & Train                                & Test (no-FT)                         & Test (FT)                            \\ 
HDT   & 69K             & 0.5\%          & $0.89\pm 0.00$          & $0.12\pm 0.01$          & $0.92\pm 0.10$          \\
PDT   & 6K              & 0.05\%         & $0.88\pm 0.00$          & $0.06\pm 0.05$          & $0.09\pm 0.01$          \\
\name  & 147K            & 0.7\%          & $\mathbf{0.92\pm 0.00}$          & $\mathbf{0.20\pm 0.01}$ & $\mathbf{0.95\pm 0.10}$  \\ \hline
\end{tabular}
\caption{Evaluation results on Meta-World ML45 benchmarks}
\label{tab:meta45}
\end{table*}


\subsection{\name improves training performance.}
\label{subsec:imp_train}

In this section, we evaluate whether adding the memory module helps improve the pre-training performance.
We report relative improvement for clear comparison: 
$\text{rel-imp(\%)} = \frac{(\text{model score} - \text{best score in data)}}{\text{best score in data}}\times 100$ to measure the model performance. For better visualization, we take the logarithm of the rel-imp(\%). As shown in Figure~\ref{fig:adapt} (bottom), the proposed \name out performs MDT in 13 out of 17 games.
\name outperforms RMDT in 15 out of 17 games.
These results demonstrate that the memory module improves the policy training performance.

\section{Conclusion}
LLM-based RL algorithms have shown generalization across multiple tasks and games. 
We argue that this ability comes from implicit memory that fits a large number of parameters to the training data, which is inefficient in terms of model size.
In contrast, we propose a new approach inspired by the concept of ``working memory'' called \fullname (\name), which stores training experience explicitly in a content-addressable matrix module for later retrieval and use. 
The evaluation demonstrates that \name achieves better generalization on Atari games with only 10\% of the model parameters compared to the state-of-the-art method. 
We also show that \name outperforms other memory-based DT methods regarding generalization and adaptability.
Furthermore, we demonstrate that fine-tuning \name with a small amount of data can produce state-of-the-art results on both Atari games and the Meta-World environment, when compared to MDT, RMDT, PDT, and HDT. 

\section*{Impact Statement}
\label{section:broader_impact}
We do not foresee any significant societal impact resulting from our proposed method. The current algorithm is not yet designed to interact with humans or any realistic environment. It is primarily developed within a controlled and theoretical framework, focusing on improving computational techniques rather than practical applications at this stage.
If one chooses to extend our methods to more interactive or real-world scenarios, caution should be exercised to ensure that any safety and ethical concerns are appropriately addressed. 
As our work is categorized in the offline-RL domain, it is feasible to supplement its training with a dataset that aligns with human intents and values. This alignment could help in creating systems that are more beneficial and less harmful when eventually deployed in practical applications. However, one must be wary that the way our architecture generalizes across tasks is still not well understood, and as a consequence, we cannot guarantee the generalization of its desirable features: performance, robustness, fairness, etc.
Moreover, we contribute to increasing large models accessibility and reducing their ecological impact. Efficient algorithms can lower the computational resources required, leading to less energy consumption and a smaller carbon footprint. 
In summary, while our current work is confined to theoretical advancements, the potential extension to practical applications must be approached with diligence. Ensuring safety, ethical integrity, and sustainability will be paramount as these methods evolve from offline settings to more dynamic and human-centric environments.

\bibliography{ref}

\begin{thebibliography}{47}
\providecommand{\natexlab}[1]{#1}
\providecommand{\url}[1]{\texttt{#1}}
\expandafter\ifx\csname urlstyle\endcsname\relax
  \providecommand{\doi}[1]{doi: #1}\else
  \providecommand{\doi}{doi: \begingroup \urlstyle{rm}\Url}\fi

\bibitem[Ba et~al.(2016)Ba, Hinton, Mnih, Leibo, and Ionescu]{ba2016using}
Ba, J., Hinton, G.~E., Mnih, V., Leibo, J.~Z., and Ionescu, C.
\newblock Using fast weights to attend to the recent past.
\newblock \emph{Advances in neural information processing systems}, 29, 2016.

\bibitem[Baddeley(2003)]{baddeley2003working}
Baddeley, A.
\newblock Working memory: looking back and looking forward.
\newblock \emph{Nature reviews neuroscience}, 4\penalty0 (10):\penalty0 829--839, 2003.

\bibitem[Bessonov et~al.(2023)Bessonov, Staroverov, Zhang, Kovalev, Yudin, and Panov]{bessonov2023recurrent}
Bessonov, A., Staroverov, A., Zhang, H., Kovalev, A.~K., Yudin, D., and Panov, A.~I.
\newblock Recurrent memory decision transformer.
\newblock \emph{arXiv preprint arXiv:2306.09459}, 2023.

\bibitem[Borgeaud et~al.(2022)Borgeaud, Mensch, Hoffmann, Cai, Rutherford, Millican, van~den Driessche, Lespiau, Damoc, Clark, de~Las~Casas, Guy, Menick, Ring, Hennigan, Huang, Maggiore, Jones, Cassirer, Brock, Paganini, Irving, Vinyals, Osindero, Simonyan, Rae, Elsen, and Sifre]{DBLP:conf/icml/BorgeaudMHCRM0L22}
Borgeaud, S., Mensch, A., Hoffmann, J., Cai, T., Rutherford, E., Millican, K., van~den Driessche, G., Lespiau, J., Damoc, B., Clark, A., de~Las~Casas, D., Guy, A., Menick, J., Ring, R., Hennigan, T., Huang, S., Maggiore, L., Jones, C., Cassirer, A., Brock, A., Paganini, M., Irving, G., Vinyals, O., Osindero, S., Simonyan, K., Rae, J.~W., Elsen, E., and Sifre, L.
\newblock Improving language models by retrieving from trillions of tokens.
\newblock In \emph{{ICML}}, volume 162 of \emph{Proceedings of Machine Learning Research}, pp.\  2206--2240. {PMLR}, 2022.

\bibitem[Brandfonbrener et~al.(2022)Brandfonbrener, Bietti, Buckman, Laroche, and Bruna]{DBLP:conf/nips/BrandfonbrenerB22}
Brandfonbrener, D., Bietti, A., Buckman, J., Laroche, R., and Bruna, J.
\newblock When does return-conditioned supervised learning work for offline reinforcement learning?
\newblock In \emph{NeurIPS}, 2022.

\bibitem[Brown et~al.(2020)Brown, Mann, Ryder, Subbiah, Kaplan, Dhariwal, Neelakantan, Shyam, Sastry, Askell, Agarwal, Herbert{-}Voss, Krueger, Henighan, Child, Ramesh, Ziegler, Wu, Winter, Hesse, Chen, Sigler, Litwin, Gray, Chess, Clark, Berner, McCandlish, Radford, Sutskever, and Amodei]{DBLP:journals/corr/abs-2005-14165}
Brown, T.~B., Mann, B., Ryder, N., Subbiah, M., Kaplan, J., Dhariwal, P., Neelakantan, A., Shyam, P., Sastry, G., Askell, A., Agarwal, S., Herbert{-}Voss, A., Krueger, G., Henighan, T., Child, R., Ramesh, A., Ziegler, D.~M., Wu, J., Winter, C., Hesse, C., Chen, M., Sigler, E., Litwin, M., Gray, S., Chess, B., Clark, J., Berner, C., McCandlish, S., Radford, A., Sutskever, I., and Amodei, D.
\newblock Language models are few-shot learners.
\newblock \emph{CoRR}, abs/2005.14165, 2020.

\bibitem[Chen et~al.(2022)Chen, Chen, Ma, Liu, and Liu]{DBLP:CanBi}
Chen, C., Chen, X., Ma, C., Liu, Z., and Liu, X.
\newblock Gradient-based bi-level optimization for deep learning: {A} survey.
\newblock \emph{CoRR}, abs/2207.11719, 2022.

\bibitem[Chen et~al.(2021)Chen, Lu, Rajeswaran, Lee, Grover, Laskin, Abbeel, Srinivas, and Mordatch]{DBLP:conf/nips/ChenLRLGLASM21}
Chen, L., Lu, K., Rajeswaran, A., Lee, K., Grover, A., Laskin, M., Abbeel, P., Srinivas, A., and Mordatch, I.
\newblock Decision transformer: Reinforcement learning via sequence modeling.
\newblock In \emph{NeurIPS}, pp.\  15084--15097, 2021.

\bibitem[Clark et~al.(2022)Clark, de~Las~Casas, Guy, Mensch, Paganini, Hoffmann, Damoc, Hechtman, Cai, Borgeaud, van~den Driessche, Rutherford, Hennigan, Johnson, Cassirer, Jones, Buchatskaya, Budden, Sifre, Osindero, Vinyals, Ranzato, Rae, Elsen, Kavukcuoglu, and Simonyan]{DBLP:conf/icml/ClarkCGMPHDHCB022}
Clark, A., de~Las~Casas, D., Guy, A., Mensch, A., Paganini, M., Hoffmann, J., Damoc, B., Hechtman, B.~A., Cai, T., Borgeaud, S., van~den Driessche, G., Rutherford, E., Hennigan, T., Johnson, M.~J., Cassirer, A., Jones, C., Buchatskaya, E., Budden, D., Sifre, L., Osindero, S., Vinyals, O., Ranzato, M., Rae, J.~W., Elsen, E., Kavukcuoglu, K., and Simonyan, K.
\newblock Unified scaling laws for routed language models.
\newblock In \emph{{ICML}}, volume 162 of \emph{Proceedings of Machine Learning Research}, pp.\  4057--4086. {PMLR}, 2022.

\bibitem[Cowan(2008)]{cowan2008differences}
Cowan, N.
\newblock What are the differences between long-term, short-term, and working memory?
\newblock \emph{Progress in brain research}, 169:\penalty0 323--338, 2008.

\bibitem[Csord{\'a}s \& Schmidhuber(2019)Csord{\'a}s and Schmidhuber]{csordas2019improving}
Csord{\'a}s, R. and Schmidhuber, J.
\newblock Improving differentiable neural computers through memory masking, de-allocation, and link distribution sharpness control.
\newblock \emph{arXiv preprint arXiv:1904.10278}, 2019.

\bibitem[Das et~al.(1992)Das, Giles, and Sun]{das1992learning}
Das, S., Giles, C.~L., and Sun, G.-Z.
\newblock Learning context-free grammars: Capabilities and limitations of a recurrent neural network with an external stack memory.
\newblock In \emph{Proceedings of The Fourteenth Annual Conference of Cognitive Science Society. Indiana University}, volume~14, 1992.

\bibitem[Devlin et~al.(2019)Devlin, Chang, Lee, and Toutanova]{DBLP:conf/naacl/DevlinCLT19}
Devlin, J., Chang, M., Lee, K., and Toutanova, K.
\newblock {BERT:} pre-training of deep bidirectional transformers for language understanding.
\newblock In \emph{{NAACL-HLT} {(1)}}, pp.\  4171--4186. Association for Computational Linguistics, 2019.

\bibitem[Dooley(2007)]{DooleyArchivalAdvantage}
Dooley, J.
\newblock The archival advantage: Integrating archival expertise into management of born-digital library materials.
\newblock \emph{Archival Science Special Issue on Archiving Research Data}, 7\penalty0 (1), March 2007.

\bibitem[Dosovitskiy et~al.(2021)Dosovitskiy, Beyer, Kolesnikov, Weissenborn, Zhai, Unterthiner, Dehghani, Minderer, Heigold, Gelly, Uszkoreit, and Houlsby]{DBLP:conf/iclr/DosovitskiyB0WZ21}
Dosovitskiy, A., Beyer, L., Kolesnikov, A., Weissenborn, D., Zhai, X., Unterthiner, T., Dehghani, M., Minderer, M., Heigold, G., Gelly, S., Uszkoreit, J., and Houlsby, N.
\newblock An image is worth 16x16 words: Transformers for image recognition at scale.
\newblock In \emph{{ICLR}}. OpenReview.net, 2021.

\bibitem[Eslami et~al.(2016)Eslami, Heess, Weber, Tassa, Szepesvari, Kavukcuoglu, and Hinton]{DBLP:conf/nips/EslamiHWTSKH16}
Eslami, S. M.~A., Heess, N., Weber, T., Tassa, Y., Szepesvari, D., Kavukcuoglu, K., and Hinton, G.~E.
\newblock Attend, infer, repeat: Fast scene understanding with generative models.
\newblock In \emph{{NIPS}}, pp.\  3225--3233, 2016.

\bibitem[Fan et~al.(2021)Fan, Gardent, Braud, and Bordes]{fan2021augmenting}
Fan, A., Gardent, C., Braud, C., and Bordes, A.
\newblock Augmenting transformers with knn-based composite memory for dialog.
\newblock \emph{Transactions of the Association for Computational Linguistics}, 9:\penalty0 82--99, 2021.

\bibitem[Goldman-Rakic(1995)]{goldman1995cellular}
Goldman-Rakic, P.~S.
\newblock Cellular basis of working memory.
\newblock \emph{Neuron}, 14\penalty0 (3):\penalty0 477--485, 1995.

\bibitem[Goyal et~al.(2022)Goyal, Didolkar, Lamb, Badola, Ke, Rahaman, Binas, Blundell, Mozer, and Bengio]{DBLP:conf/iclr/GoyalDLBKRBBMB22}
Goyal, A., Didolkar, A.~R., Lamb, A., Badola, K., Ke, N.~R., Rahaman, N., Binas, J., Blundell, C., Mozer, M.~C., and Bengio, Y.
\newblock Coordination among neural modules through a shared global workspace.
\newblock In \emph{{ICLR}}. OpenReview.net, 2022.

\bibitem[Graves et~al.(2014)Graves, Wayne, and Danihelka]{graves2014neural}
Graves, A., Wayne, G., and Danihelka, I.
\newblock Neural turing machines.
\newblock \emph{arXiv preprint arXiv:1410.5401}, 2014.

\bibitem[Hochreiter \& Schmidhuber(1997)Hochreiter and Schmidhuber]{hochreiter1997long}
Hochreiter, S. and Schmidhuber, J.
\newblock Long short-term memory.
\newblock \emph{Neural computation}, 9\penalty0 (8):\penalty0 1735--1780, 1997.

\bibitem[Hu et~al.(2022)Hu, Shen, Wallis, Allen{-}Zhu, Li, Wang, Wang, and Chen]{DBLP:conf/iclr/HuSWALWWC22}
Hu, E.~J., Shen, Y., Wallis, P., Allen{-}Zhu, Z., Li, Y., Wang, S., Wang, L., and Chen, W.
\newblock Lora: Low-rank adaptation of large language models.
\newblock In \emph{{ICLR}}. OpenReview.net, 2022.

\bibitem[Humphreys et~al.(2022)Humphreys, Guez, Tieleman, Sifre, Weber, and Lillicrap]{DBLP:conf/nips/HumphreysGTSWL22}
Humphreys, P.~C., Guez, A., Tieleman, O., Sifre, L., Weber, T., and Lillicrap, T.~P.
\newblock Large-scale retrieval for reinforcement learning.
\newblock In \emph{NeurIPS}, 2022.

\bibitem[Janner et~al.(2021)Janner, Li, and Levine]{janner2021sequence}
Janner, M., Li, Q., and Levine, S.
\newblock Offline reinforcement learning as one big sequence modeling problem.
\newblock In \emph{Advances in Neural Information Processing Systems}, 2021.

\bibitem[Kaplan et~al.(2020)Kaplan, McCandlish, Henighan, Brown, Chess, Child, Gray, Radford, Wu, and Amodei]{kaplan2020scaling}
Kaplan, J., McCandlish, S., Henighan, T., Brown, T.~B., Chess, B., Child, R., Gray, S., Radford, A., Wu, J., and Amodei, D.
\newblock Scaling laws for neural language models.
\newblock \emph{arXiv preprint arXiv:2001.08361}, 2020.

\bibitem[Lee et~al.(2022)Lee, Nachum, Yang, Lee, Freeman, Guadarrama, Fischer, Xu, Jang, Michalewski, and Mordatch]{DBLP:conf/nips/LeeNYLFGFXJMM22}
Lee, K., Nachum, O., Yang, M., Lee, L., Freeman, D., Guadarrama, S., Fischer, I., Xu, W., Jang, E., Michalewski, H., and Mordatch, I.
\newblock Multi-game decision transformers.
\newblock In \emph{NeurIPS}, 2022.

\bibitem[Martins et~al.(2022)Martins, Marinho, and Martins]{DBLP:conf/acl/MartinsMM22}
Martins, P.~H., Marinho, Z., and Martins, A. F.~T.
\newblock {\(\infty\)}-former: Infinite memory transformer.
\newblock In \emph{{ACL} {(1)}}, pp.\  5468--5485. Association for Computational Linguistics, 2022.

\bibitem[Munkhdalai \& Yu(2017)Munkhdalai and Yu]{munkhdalai2017meta}
Munkhdalai, T. and Yu, H.
\newblock Meta networks.
\newblock In \emph{International conference on machine learning}, pp.\  2554--2563. PMLR, 2017.

\bibitem[Munkhdalai et~al.(2018)Munkhdalai, Yuan, Mehri, and Trischler]{pmlr-v80-munkhdalai18a}
Munkhdalai, T., Yuan, X., Mehri, S., and Trischler, A.
\newblock Rapid adaptation with conditionally shifted neurons.
\newblock In Dy, J. and Krause, A. (eds.), \emph{Proceedings of the 35th International Conference on Machine Learning}, volume~80 of \emph{Proceedings of Machine Learning Research}, pp.\  3664--3673. PMLR, 10--15 Jul 2018.
\newblock URL \url{https://proceedings.mlr.press/v80/munkhdalai18a.html}.

\bibitem[Munkhdalai et~al.(2019)Munkhdalai, Sordoni, Wang, and Trischler]{munkhdalai2019metalearned}
Munkhdalai, T., Sordoni, A., Wang, T., and Trischler, A.
\newblock Metalearned neural memory.
\newblock \emph{Advances in Neural Information Processing Systems}, 32, 2019.

\bibitem[Neyshabur et~al.(2019)Neyshabur, Li, Bhojanapalli, LeCun, and Srebro]{DBLP:conf/iclr/NeyshaburLBLS19}
Neyshabur, B., Li, Z., Bhojanapalli, S., LeCun, Y., and Srebro, N.
\newblock The role of over-parametrization in generalization of neural networks.
\newblock In \emph{{ICLR} (Poster)}. OpenReview.net, 2019.

\bibitem[OpenAI(2023)]{DBLP:journals/corr/abs-2303-08774}
OpenAI.
\newblock {GPT-4} technical report.
\newblock \emph{CoRR}, abs/2303.08774, 2023.

\bibitem[Radford et~al.(2019)Radford, Wu, Child, Luan, Amodei, Sutskever, et~al.]{radford2019language}
Radford, A., Wu, J., Child, R., Luan, D., Amodei, D., Sutskever, I., et~al.
\newblock Language models are unsupervised multitask learners.
\newblock \emph{OpenAI blog}, 1\penalty0 (8):\penalty0 9, 2019.

\bibitem[Ramsauer et~al.(2020)Ramsauer, Sch{\"a}fl, Lehner, Seidl, Widrich, Adler, Gruber, Holzleitner, Pavlovi{\'c}, Sandve, et~al.]{ramsauer2020hopfield}
Ramsauer, H., Sch{\"a}fl, B., Lehner, J., Seidl, P., Widrich, M., Adler, T., Gruber, L., Holzleitner, M., Pavlovi{\'c}, M., Sandve, G.~K., et~al.
\newblock Hopfield networks is all you need.
\newblock \emph{arXiv preprint arXiv:2008.02217}, 2020.

\bibitem[Santoro et~al.(2016)Santoro, Bartunov, Botvinick, Wierstra, and Lillicrap]{santoro2016meta}
Santoro, A., Bartunov, S., Botvinick, M., Wierstra, D., and Lillicrap, T.
\newblock Meta-learning with memory-augmented neural networks.
\newblock In \emph{International conference on machine learning}, pp.\  1842--1850. PMLR, 2016.

\bibitem[Schmidhuber(1992)]{schmidhuber1992learning}
Schmidhuber, J.
\newblock Learning to control fast-weight memories: An alternative to dynamic recurrent networks.
\newblock \emph{Neural Computation}, 4\penalty0 (1):\penalty0 131--139, 1992.

\bibitem[Sukhbaatar et~al.(2015)Sukhbaatar, Weston, Fergus, et~al.]{sukhbaatar2015end}
Sukhbaatar, S., Weston, J., Fergus, R., et~al.
\newblock End-to-end memory networks.
\newblock \emph{Advances in neural information processing systems}, 28, 2015.

\bibitem[Touvron et~al.(2023)Touvron, Lavril, Izacard, Martinet, Lachaux, Lacroix, Rozi{\`{e}}re, Goyal, Hambro, Azhar, Rodriguez, Joulin, Grave, and Lample]{DBLP:journals/corr/abs-2302-13971}
Touvron, H., Lavril, T., Izacard, G., Martinet, X., Lachaux, M., Lacroix, T., Rozi{\`{e}}re, B., Goyal, N., Hambro, E., Azhar, F., Rodriguez, A., Joulin, A., Grave, E., and Lample, G.
\newblock Llama: Open and efficient foundation language models.
\newblock \emph{CoRR}, abs/2302.13971, 2023.

\bibitem[Vaswani et~al.(2017)Vaswani, Shazeer, Parmar, Uszkoreit, Jones, Gomez, Kaiser, and Polosukhin]{DBLP:conf/nips/VaswaniSPUJGKP17}
Vaswani, A., Shazeer, N., Parmar, N., Uszkoreit, J., Jones, L., Gomez, A.~N., Kaiser, L., and Polosukhin, I.
\newblock Attention is all you need.
\newblock In \emph{{NIPS}}, pp.\  5998--6008, 2017.

\bibitem[von Oswald et~al.(2020)von Oswald, Henning, Sacramento, and Grewe]{DBLP:conf/iclr/OswaldHSG20}
von Oswald, J., Henning, C., Sacramento, J., and Grewe, B.~F.
\newblock Continual learning with hypernetworks.
\newblock In \emph{{ICLR}}. OpenReview.net, 2020.

\bibitem[Wang et~al.(2024)Wang, Dong, Cheng, Liu, Yan, Gao, and Wei]{wang2024augmenting}
Wang, W., Dong, L., Cheng, H., Liu, X., Yan, X., Gao, J., and Wei, F.
\newblock Augmenting language models with long-term memory.
\newblock \emph{Advances in Neural Information Processing Systems}, 36, 2024.

\bibitem[Wu et~al.(2022{\natexlab{a}})Wu, Li, Mangalam, Fan, Xiong, Malik, and Feichtenhofer]{wu2022memvit}
Wu, C.-Y., Li, Y., Mangalam, K., Fan, H., Xiong, B., Malik, J., and Feichtenhofer, C.
\newblock Memvit: Memory-augmented multiscale vision transformer for efficient long-term video recognition.
\newblock In \emph{Proceedings of the IEEE/CVF Conference on Computer Vision and Pattern Recognition}, pp.\  13587--13597, 2022{\natexlab{a}}.

\bibitem[Wu et~al.(2022{\natexlab{b}})Wu, Rabe, Hutchins, and Szegedy]{wu2022memorizing}
Wu, Y., Rabe, M.~N., Hutchins, D., and Szegedy, C.
\newblock Memorizing transformers.
\newblock \emph{arXiv preprint arXiv:2203.08913}, 2022{\natexlab{b}}.

\bibitem[Xu et~al.(2022)Xu, Shen, Zhang, Lu, Zhao, Tenenbaum, and Gan]{DBLP:conf/icml/XuSZLZTG22}
Xu, M., Shen, Y., Zhang, S., Lu, Y., Zhao, D., Tenenbaum, J.~B., and Gan, C.
\newblock Prompting decision transformer for few-shot policy generalization.
\newblock In \emph{{ICML}}, volume 162 of \emph{Proceedings of Machine Learning Research}, pp.\  24631--24645. {PMLR}, 2022.

\bibitem[Xu et~al.(2023)Xu, Lu, Shen, Zhang, Zhao, and Gan]{DBLP:journals/corr/abs-2304-08487}
Xu, M., Lu, Y., Shen, Y., Zhang, S., Zhao, D., and Gan, C.
\newblock Hyper-decision transformer for efficient online policy adaptation.
\newblock \emph{CoRR}, abs/2304.08487, 2023.

\bibitem[Yu et~al.(2019)Yu, Quillen, He, Julian, Hausman, Finn, and Levine]{yu2019meta}
Yu, T., Quillen, D., He, Z., Julian, R., Hausman, K., Finn, C., and Levine, S.
\newblock Meta-world: A benchmark and evaluation for multi-task and meta reinforcement learning.
\newblock In \emph{Conference on Robot Learning (CoRL)}, 2019.
\newblock URL \url{https://arxiv.org/abs/1910.10897}.

\bibitem[Zheng et~al.(2022)Zheng, Zhang, and Grover]{DBLP:conf/icml/ZhengZG22}
Zheng, Q., Zhang, A., and Grover, A.
\newblock Online decision transformer.
\newblock In \emph{{ICML}}, volume 162 of \emph{Proceedings of Machine Learning Research}, pp.\  27042--27059. {PMLR}, 2022.

\end{thebibliography}
\bibliographystyle{icml2024}

\newpage
\appendix
\onecolumn
\section{Implementation Details}



\subsection{Hyper-parameters}

This section will delve into the specifics of the model parameters. Understanding these parameters is key to understanding the model's workings.
It is worth noting that the source code for this model is publicly available at \url{https://anonymous.4open.science/r/DT-Mem-Submission277/README.md}. 
This allows for a deeper understanding of the model's inner workings and may facilitate replicating its results.

\begin{table}[htbp]
\centering
\begin{tabular}{|c|c|}
\hline
Hyperparameters             & Value    \\ \hline
K (length of context)       & 28       \\ \hline
dropout rate                & 0.1      \\ \hline
maximum epochs              & 1000     \\ \hline
steps for each epoch        & 1000     \\ \hline
optimizer learning rate     & 1e-4     \\ \hline
weight decay                & 1e-4     \\ \hline
gradient norm clip          & 1.       \\ \hline
data points for each dataset & 500,000  \\ \hline
batch size                  & 64       \\ \hline
memory slots                & 1290     \\ \hline
activation                  & GELU     \\ \hline
optimizer                   & AdamW    \\ \hline
scheduler                   & LambdaLR \\ \hline
\end{tabular}
\caption{Hyperparameters for \name training}
\end{table}


\subsection{Training and fine-tuning algorithm}
\label{subsec:train_algo}

In this section, we present the pre-training \name in Appendix \ref{subsec:train_algo} and fine-tuning \name with LoRA in Appendix \ref{subsec:fine-tune}.

We pre-train \name on multiple offline datasets.
Each gradient update of the \name model considers information from each training task.

\begin{algorithm}[htbp]
\caption{Pre-train \name}
\label{alg:alg1}
\begin{algorithmic}[1]
\FOR{T episodes}
\FOR{Task $\mathcal{T}_i\in T^{\text{train}}$}
\STATE Sample trajectories $\tau = (s_0,a_0,r_0,\cdots,s_H,a_H,r_H)$ from the dataset $\mathcal{D}_i$. 
\STATE Split trajectories into different segments with length K and calculate return-to-go in the input sequence.
\STATE Given $\hat{\tau}_{t+1:t+K}$, compute the sequence embedding $e_{\text{seq}}$.
\STATE Update the memory module and retrieve the relative information as $\bm{E}_{\text{out}}$
\STATE Given $\bm{E}_{\text{out}}$, predict actions $\Tilde{a}_t$, reward $\Tilde{r}_t$, and return-to-go $\Tilde{R}_t$.
\STATE Compute the loss according to Eqn.~\ref{eqn:loss}.
\STATE Update all module parameters.
\ENDFOR
\ENDFOR
\end{algorithmic}
\end{algorithm}

We fine-tune the memory module to adapt to each downstream task. 
To achieve this, we fix the pre-trained \name model parameters and add additional LoRA parameters for the memory module feed-forward neural networks. 
The fine-tuning dataset is used to update these LoRA parameters only.

\begin{algorithm}[htbp]
\caption{Fine-tuning \name}
\label{alg:alg2}
\textbf{Require:} Fine-tuning dataset $\mathcal{T}^i\in T^{\text{test}}$ dataset $\mathcal{D}^i$ for $\mathcal{T}^i$. Initialize LoRA parameters $\bm{\hat{B}}^q, \bm{\hat{B}}^k, \bm{\hat{B}}^v, \bm{\hat{A}}^q, \bm{\hat{A}}^k, \bm{\hat{A}}^v, \bm{B}^q, \bm{A}^q, \bm{B}^k, \bm{A}^k$.
\begin{algorithmic}[1]
\FOR{T steps}
\STATE Split trajectories into different segments with length K and calculate return-to-go in the input sequence.
\STATE Given $\hat{\tau}_{t+1:t+K}$, compute the sequence embedding $e_{\text{seq}}$.
\STATE Update memory module using $\bm{\hat{Q}}=\bm{M}(\bm{\hat{W}}^q+\bm{\hat{B}}^q\bm{\hat{A}}^q)$,
$\bm{\hat{K}}=\bm{M}(\bm{\hat{W}}^k+\bm{\hat{B}}^k\bm{\hat{A}}^k)$,$\bm{\hat{V}}=\bm{M}(\bm{\hat{W}}^v+\bm{\hat{B}}^v\bm{\hat{A}}^v)$,
$\bm{Q}=\bm{M}(\bm{W}^q+\bm{B}^q\bm{A}^q)$,$\bm{K}=\bm{M}(\bm{W}^k+\bm{B}^k\bm{A}^k)$
\STATE Retrieve the relative information as $\bm{E}_{\text{out}}$
\STATE Given $\bm{E}_{\text{out}}$, predict actions $\Tilde{a}_t$, reward $\Tilde{r}_t$, and return-to-go $\Tilde{R}_t$.
\STATE Compute the loss according to Eqn.~\ref{eqn:loss}.
\STATE Update LoRA parameters only.
\ENDFOR
\end{algorithmic}
\end{algorithm}

\begin{algorithm}
\caption{Memory Operations}
\begin{algorithmic}[1]
\STATE \textbf{Step 0: Memory Module Initialization}
\STATE Initialize memory as a random matrix $\bm{M}$ where each row $\bm{m_i} \in \mathbb{R}^d$ and $i \in [0, N]$.
\STATE
\STATE \textbf{Step 1: Input Sequence Organizing}
\STATE Restructure input sequence into format \(<\hat{r}_t, s_t, a_t>\).
\STATE Define embedding functions \(g_s(s) = e_s\), \(g_a(a) = e_a\), \(g_r(\hat{r}) = e_{\hat{r}}\).
\STATE Concatenate embeddings to form input sequence $\bm{E} = [\cdots; \bm{e}_{s_t}, \bm{e}_{a_t}, \bm{e}_{\hat{r}_t}; \cdots]$.
\STATE
\STATE \textbf{Step 2: Content-based Address}
\STATE Use attention to locate the memory slot for new input.
\STATE Calculate position address $\bm{w} = \text{softmax}\Big(\frac{\bm{Q}\bm{K}^T}{\sqrt{d}}\Big)$.
\STATE Define $\bm{Q} = \bm{M}\bm{W}^q$ and $\bm{K} = \bm{E}\bm{W}^k$.
\STATE
\FOR{N Times memory operations do}
\STATE \textbf{Step 3: Memory Update}
\STATE Calculate erasing vector $\bm{\epsilon}^e$ and adding vector $\bm{\epsilon}^a$.
\STATE Define $\bm{\hat{Q}} = \bm{M}\bm{\hat{W}}^q$, $\bm{\hat{K}} = \bm{E}\bm{\hat{W}}^k$, $\bm{\hat{V}} = \bm{E}\bm{\hat{W}}^v$.
\STATE Compute writing strength $\beta = \text{softmax}\Big(\frac{\bm{\hat{Q}}\bm{\hat{K}}^T}{\sqrt{d}}\Big)$.
\STATE Calculate $\bm{\epsilon}^e = \bm{w} \odot (1 - \beta)$.
\STATE Calculate $\bm{\epsilon}^a = (\bm{w} \odot \beta) \bm{\hat{W}}^v \bm{x}$.
\STATE Update memory $\bm{M}_n = \bm{M}_{n-1} \odot (\bm{1} - \bm{\epsilon}^e) + \bm{\epsilon}^a$.
\STATE
\STATE \textbf{Step 4: Memory Retrieve}
\STATE Retrieve information from memory for decision-making.
\STATE Compute read position vector using content-based address.
\STATE Retrieve memory $\bm{E}_{\text{out}} = \bm{w} \odot \bm{M}_n$.
\STATE $\bm{E} = \bm{E}_{\text{out}}$
\ENDFOR
\STATE output $\bm{E}$ for action decoder.
\end{algorithmic}
\end{algorithm}

\section{Additional Experiments}

\subsection{Evaluation Parameters}

To evaluate the performance of our model on Atari games, we randomly selected 16 different random seeds for evaluation. 
We chose the random seed by multiplying the number of runs by 100. 
For example, the random seed for run 6 is $6 \times 100 = 600$.




\subsection{Training Efficiencies}
\label{subsec:train_eff}

\begin{figure}[htbp]
     \centering
     \begin{subfigure}
         \centering
         \includegraphics[width=0.48\textwidth]{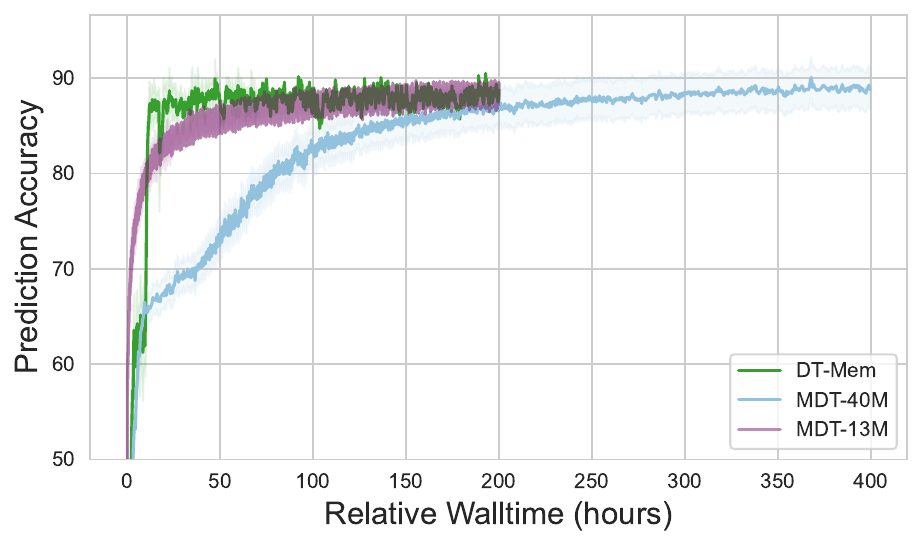}
     \end{subfigure}
     \hfill
     \begin{subfigure}
         \centering
         \includegraphics[width=0.48\textwidth]{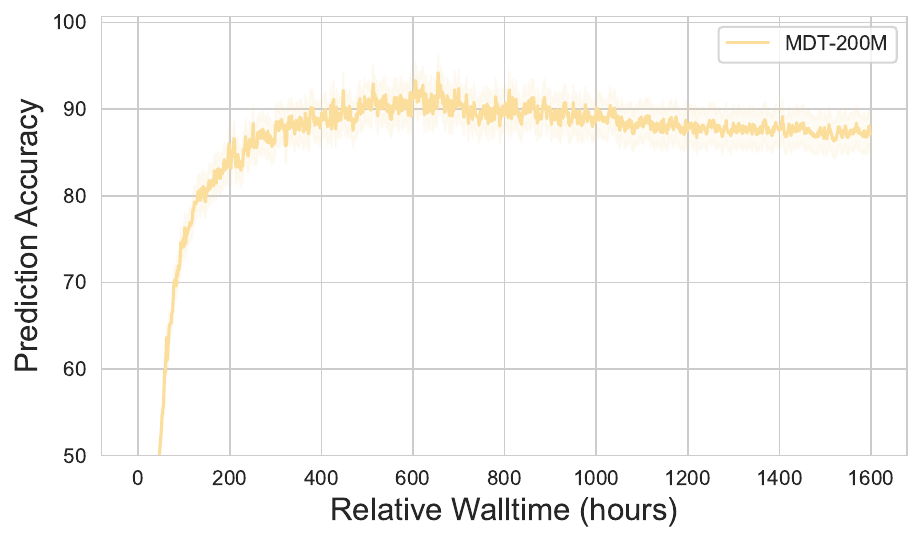}
     \end{subfigure}
\caption{This graph shows the prediction accuracy during training. Each curve represents three runs with different random seeds. For better visualization, MDT-200M is displayed in a separate figure.}
\label{fig:train_eff}
\end{figure}

We illustrate the model training curve in Figure \ref{fig:train_eff} to demonstrate training efficiency.
It is reasonable to report the prediction loss on the training dataset for the training curve since we use a supervised loss.
Here, the prediction accuracy consists of three parts: action prediction accuracy, reward prediction accuracy, and return prediction accuracy.
The y-axis shows the average value of these three predictions, and the x-axis is the relative walltime based on the same computing resources.

\subsection{The analysis of memory size}

\begin{figure}[htbp]
    \centering
    \includegraphics[width=0.8\textwidth]{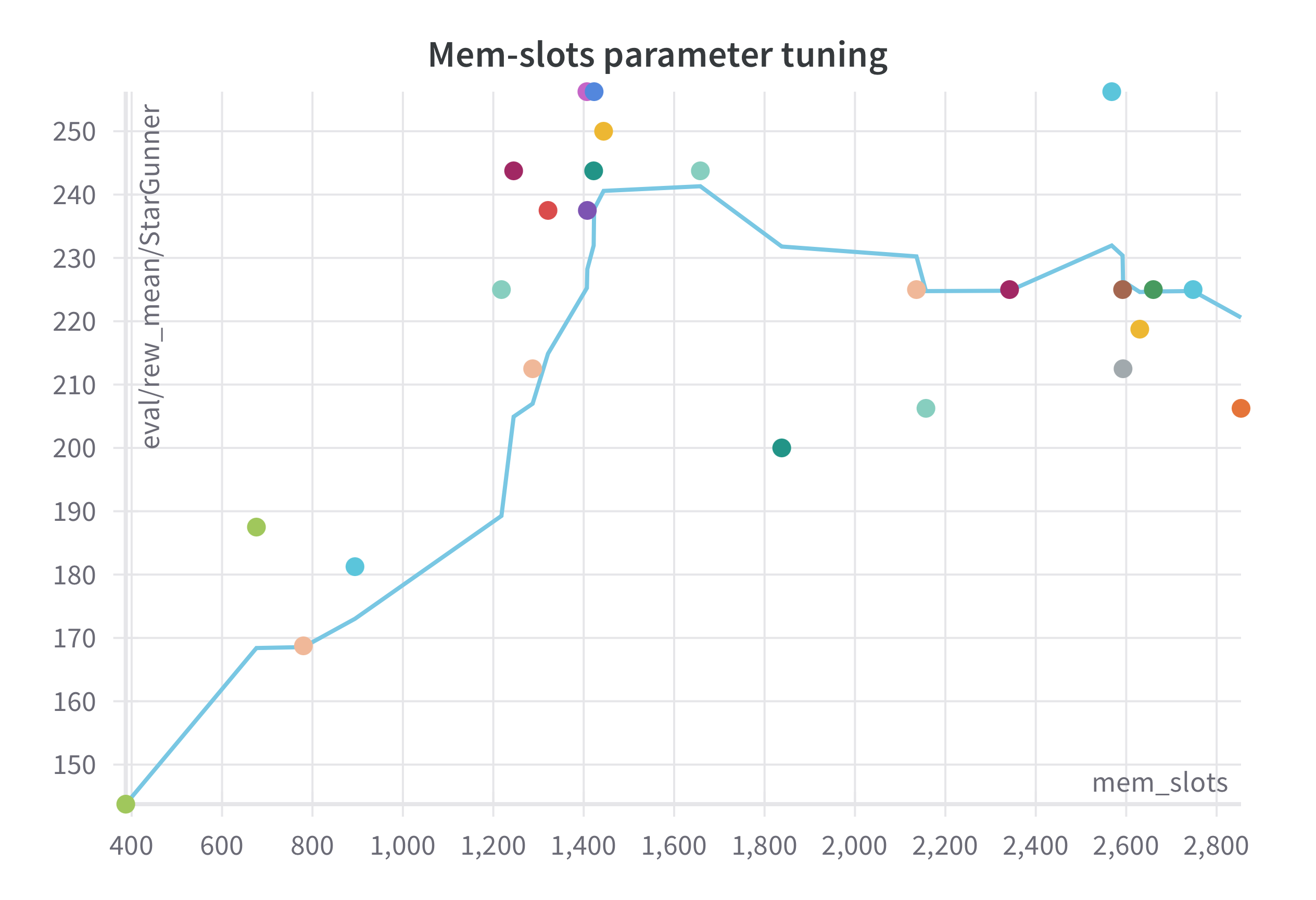}
    \caption{The parameter tuning results for the number of memory slots. The blue curve shows the like from left to right over the x-axis and plots the running average y value.}
    \label{fig:mem_slot_tuning}
\end{figure}

In this section, we investigate the impact of the memory module size on the performance of \name. 
We employ the Bayes optimization strategy to tune the parameters. 
It's worth noting that the memory size is calculated by multiplying the number of memory slots by the size of each slot, which is fixed at 512 dimensions for the sake of evaluation simplicity. 
To expedite the hyper-parameter tuning process, we present the evaluation results based on 100k training steps of the StarGunner game.
We assess various configurations of memory slots and calculate their corresponding average rewards over 16 runs. 
Figure~\ref{fig:mem_slot_tuning} reveals several key findings:
(1)Increasing the size of memory slots leads to a higher reward accumulation. 
Notably, there is a significant performance boost when the number exceeds 1200.
(2)In summary, when the number of memory slots exceeds 1800, the system's performance decreases. This decline occurs because there is a trade-off between the number of memory slots and the training steps. With a larger number of memory slots, it becomes necessary to allocate more training time.

\subsection{Ablation study of LoRA adaptor}

\begin{table}[htbp]
\centering
\begin{tabular}{c|ccc|c|cc}
\hline
                   & \multicolumn{3}{c|}{Meta-World ML45 Performances}                           & \multicolumn{1}{l|}{Data size} & \multicolumn{2}{c}{Model} \\
                   & Train                   & Test (no-FT)            & Test (FT)               &                                & Adap.  & Per.  \\ \hline
DT-Mem (hyper-net) & $0.92\pm 0.01$          & $\mathbf{0.23\pm 0.10}$ & $0.81\pm 0.15$          & 30                             & 5.7M        & 43.8\%      \\
DT-Mem             & $\mathbf{0.92\pm 0.00}$ & $0.20\pm 0.01$          & $\mathbf{0.95\pm 0.10}$ & 10                             & 147K        & 0.7\%       \\ \hline
\end{tabular}
\caption{Ablation study results on Meta-World ML45 benchmarks. \name (hyper-net) denotes the variation of \name, which substitutes the LoRA adaptation module with hyper-networks. Adap. stands for adaptation parameters, and Per. stands for the percentage of the original model.}
\label{tab:abl_hyper}
\end{table}

In this section, we conduct an ablation study of the LoRA-based memory adaptor.
We substitute the LoRA adaptor with hyper-networks.
Specifically, the parameters of the memory module are generated from hyper-networks.
This approach is based on \cite{DBLP:conf/iclr/OswaldHSG20}, where hyper-networks take task-related information as input and generate the corresponding networks for the downstream MLP.
We use the same approach and generate parameters that are conditioned on two types of inputs: the task embedding from the task encoder and the sequence embeddings from the Transformer module.

To generate task embeddings, we adopt the same idea from PDT \citep{DBLP:conf/icml/XuSZLZTG22}, demonstrating that a small part of trajectories can represent task-related information.
We further extend this idea to extract the task information fully.
To achieve this goal, we use a Neural Network as a task encoder.
Specifically, this task encoder is implemented as a Transformer encoder-like structure \cite{DBLP:conf/nips/VaswaniSPUJGKP17}.
We first formulate the first $i$ steps of collected trajectories $\tau_{0:i} = (s_0,a_0,r_0,\cdots,s_i,a_i,r_i)$ as a task specific information.
The task trajectory $\tau_{0:i}$ is treated as a sequence of inputs to the task encoder.
The output of the task encoder is a task embedding $e_{task}\in \mathbb{R}^d$, where $d$ is the dimension of the embedding.

Then, we concatenate the task embedding and sequence embedding $e=[e_{task};e_{\text{seq}}]$ and input them to the hyper-networks.
Specifically, we define the hyper-network as a function of $f_{\omega}(\cdot)$ parameterized by $\omega$.
The output $\Theta=f_{\omega}(e)$ is a set of parameters for the memory module.

According to the evaluation results in Table~\ref{tab:abl_hyper}, including a hyper-network in the \name model improves generalization without the need for fine-tuning. 
However, it is worth noting that the hyper-network variant of \name (hyper-net) exhibits higher variance compared to \name.
The primary reason for this higher variance is the uncertainty arising from the task information. Different task-related sequences are collected in each run, resulting in varying generated parameters for the memory module.
Regarding the task fine-tuning results, we observe that the LoRA module outperforms other methods. 
This finding indicates that fine-tuning with LoRA enhances the model's adaptability.
We hypothesize that the size of the hyper-networks model plays a role in these results. 
Fine-tuning a large model size (5.7M) with a small step-size (100k steps in our case) becomes challenging. 
In an effort to improve hyper-networks fine-tuning performance, we increased the fine-tuning dataset from 10k episodes to 30k episodes.
These findings suggest that LoRA-based fine-tuning demonstrates better data efficiency.

The motivations for using LoRA to fine-tune the model can be concluded in the following two reasons:

\citet{DBLP:conf/iclr/HuSWALWWC22} suggests that the LoRA method, in contrast to other adapters, maintains model quality without introducing inference latency or shortening input sequence length. Furthermore, it facilitates rapid task-switching in service deployments by sharing most model parameters.
Parameter-efficient fine-tuning (PEFT) refines a limited number of model parameters, preserving most of the pre-trained LLM parameters, which reduces computational and storage demands \citep{DBLP:conf/iclr/HuSWALWWC22}. This approach also addresses catastrophic forgetting [4] and has outperformed standard fine-tuning in low-data and out-of-domain situations [5]. Besides, the results of full parameter fine-tuning vs. PEFT are shown in Table~\ref{tab:game_performance}:

\begin{table}[htbp]
    \centering
    \begin{tabular}{c|ccc}
    \hline
    Game          & PEFT    & FFT-Single & FFT-All \\ \hline
    Alien         & 127.4\% & 116.8\%    & 113.9\% \\
    MsPacman      & 130.8\% & 122.8      & 77.1\%  \\
    SpaceInvaders & 100.8\% & 86.8\%     & 73.4\%  \\
    StarGunner    & 158.3\% & 55.7\%     & 40.6\%  \\ \hline
    \end{tabular}
    \caption{Performance comparison of PEFT across various games}
    \label{tab:game_performance}
\end{table}
where PEFT stands for LoRA fine-tuning for all games together, FFT-single means full-parameter fine-tuning on a single game only, FFT-All stands for full-tine-tuning on all games together. Results are DQN-normalized score.

\subsection{LoRA hyper-parameters tuning}

\begin{figure} [htbp]
    \centering
    \includegraphics[width=0.9\textwidth]{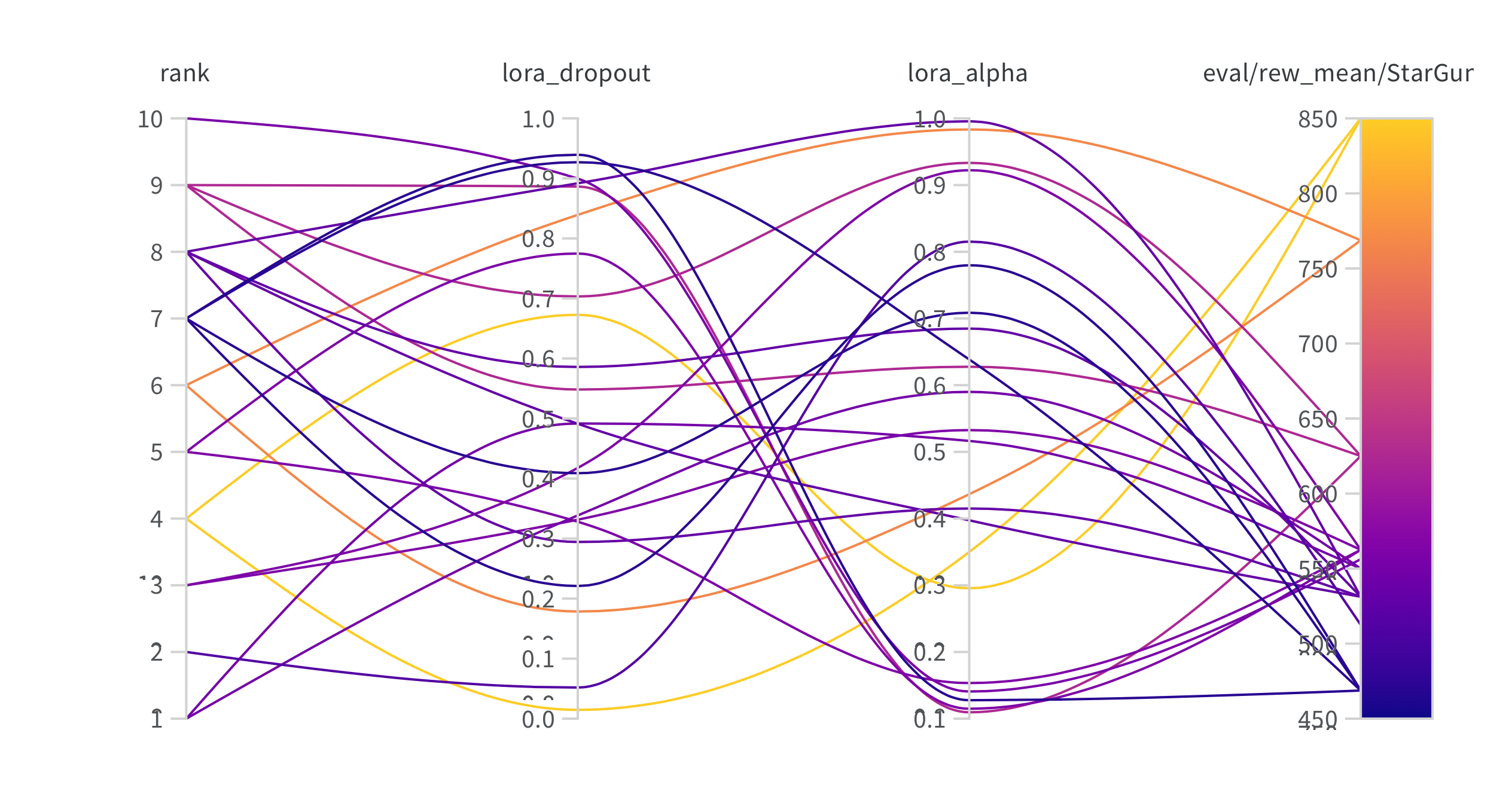}
    \caption{LoRA hyper-parameters tuning results. }
    \label{fig:lora_para}
\end{figure}

In this section, we explore the impact of LoRA hyper-parameters on the final fine-tuning results. 
LoRA employs three hyper-parameters: rank, lora\_dropout, and lora\_alpha. 
The rank parameter, denoted as $m$, determines the low-rank of adaptation matrices $\bm{B}\in\mathbb{R}^{n\times m}$ and $\bm{A}\in\mathbb{R}^{m\times d}$, as described in Section \ref{subsec:fine-tune-lora}. 
The lora\_dropout refers to the dropout rate applied to the LoRA neural networks, while lora\_alpha controls the scaling factor of the LoRA outputs.
Figure \ref{fig:lora_para} presents the fine-tuning results, with the last column (\textbf{eval/rew\_mean/StarGur}) specifically showcasing the fine-tuning results for the StarGunner game. 
To obtain the optimal set of parameters, we employ the Bayesian optimization method for parameter tuning, which suggests various parameter combinations that maximize the fine-tuning results.

\begin{table}[htbp]
\centering
\begin{tabular}{c|cc}
\hline
Parameter     & Importance score & Correlation score \\ \hline
rank          & \textbf{0.486}            & -0.132            \\ \hline
lora\_dropout & 0.285            & -0.561            \\ \hline
lora\_alpha   & 0.229            & 0.550             \\ \hline
\end{tabular}
\caption{Analysis of LoRA hyper-parameters}
\label{tab:ana_lora}
\end{table}

We further analyze these parameters and present the findings in Table \ref{tab:ana_lora}. 
To gain insights, we utilize two widely used metrics in the MLOps platform Weights\&Biases\footnote{For better understanding, please refer to \url{https://docs.wandb.ai/guides/app/features/panels/parameter-importance?_gl=1*4s7cuj*_ga*MTQxNjYxODU0OC4xNjgzNjY4Nzg3*_ga_JH1SJHJQXJ*MTY4NDc5NDkzNS40MS4xLjE2ODQ3OTQ5NDIuNTMuMC4w}}.

Regarding the \textbf{importance score}, we train a random forest model with the hyper-parameters as inputs and the metric as the target output. 
We report the feature importance values derived from the random forest. 
This hyper-parameter importance panel disentangles complex interactions among highly correlated hyper-parameters. 
It facilitates fine-tuning of hyper-parameter searches by highlighting the hyper-parameters that significantly impact the prediction of model performance.

The \textbf{correlation score} represents the linear correlation between each hyper-parameter and the chosen metric (in this case, val\_loss). 
A high correlation indicates that when the hyper-parameter has a higher value, the metric also tends to have higher values, and vice versa.
Correlation is a useful metric, but it does not capture second-order interactions between inputs and can be challenging to compare when inputs have widely different ranges.

As shown in Table \ref{tab:ana_lora}, rank emerges as the most important hyper-parameter that requires careful tuning. 
The correlation score of rank is -0.132, indicating that a smaller rank number leads to better fine-tuning results. 
Based on our findings, a rank value of 4 yields the best outcome. Lora\_dropout and lora\_alpha exhibit similar importance scores, suggesting that these two parameters can be treated equally. 
The correlation score reveals that a smaller lora\_dropout value and a larger lora\_alpha value result in improved performance.

\subsection{Ablation studies on different input sequence organizing choices}
\label{subsec:input_organize}
We examine two distinct approaches to input organization. The first approach is adopted from the trajectory Transformer as outlined in \citep{janner2021sequence}, which organizes the inputs as $(s_1,\ldots,s_t,a_1,\ldots,a_t,r_1,\ldots,r_t)$, grouping states, actions, and rewards accordingly. The second approach is derived from the decision Transformer as described in \citep{DBLP:conf/nips/ChenLRLGLASM21}, and is the method utilized in this study.

\begin{table}[ht]
\centering
\begin{tabular}{ccc}
\toprule
Game & Choice one & Choice two (Ours) \\
\midrule
Alien & 211.9 & 239.6 \\
MsPacman & 637.1 & 713.4 \\
SpaceInvaders & 165.7 & 171.2 \\
StarGunner & 620.7 & 709.3 \\
\bottomrule
\end{tabular}
\label{tab:input_organize}
\caption{Ablation studies on different choices of organizing. Each value represents raw scores in Atari games.}
\end{table}

From the table above, we observe minor differences between the two sets of inputs. However, the variance in outcomes between the two methodologies is not significant. Therefore, we empirically adopt the second approach for our design in this paper.

\subsection{Ablation studies with DT}

\begin{table}[htbp]
    \centering
    \begin{tabular}{c|ccc}
        \hline
        & DT-Mem (Ave) & DT-Mem FT (Ave) & DT-20M (Ave) \\
        \hline
        10k & - & - & 10.1\% \\
        20k & - & - & 9.8\% \\
        30k & - & - & 15.3\% \\
        40k & - & - & 22.6\% \\
        50k & 51.0\% & 127.4\% & 41.8\% \\
        100k & - & - & 83.1\% \\
        200k & - & - & 120.3\% \\
        500k & - & - & 170.7\% \\
        \hline
    \end{tabular}
    \caption{Comparison with DT in different fine-tuning datasets}
    \label{tab:ab_dt}
\end{table}

As shown in Table~\ref{tab:ab_dt}, the leftmost column represents the size of the dataset used for training. As seen in the table above, the generalized agent DT-Mem outperforms when compared to training on the DT-20M 50k datasets. Fine-tuning DT-Mem on 50k datasets yields better results than training DT-20M on 200k datasets.

\subsection{Full Fine-tuning vs. LoRA}

**Full Fine-tuning (FFT) vs. LoRA**: To assess whether the use of LoRA adversely affects performance, we conducted experiments contrasting Full Fine-Tuning (FFT) of memory parameters with LoRA. In this context, FFT-single refers to fine-tuning all parameters exclusively on a single game, whereas FFT-All represents fine-tuning simultaneously on the entire set of games. The results are DQN-normalized score.
\begin{table}[ht]
\centering
\begin{tabular}{c|c|c|c}
\hline
Game          & PEFT    & FFT-Single & FFT-All \\
\hline
Alien         & 127.4\% & 116.8\%    & 113.9\% \\
MsPacman      & 130.8\% & 122.8      & 77.1\%  \\
SpaceInvaders & 100.8\% & 86.8\%     & 73.4\%  \\
StarGunner    & 158.3\% & 55.7\%     & 40.6\% \\ \hline
\end{tabular}
\end{table}
Based on above results, we conclude the following observations:

- LoRA appears to be the most consistently effective strategy across the games provided.
- While **FFT-Single** occasionally outperforms PEFT (like in Alien), **FFT-All** consistently trails behind the other two.

The reason full fine-tuning is not comparable to PEFT comes from the following parts: 1. Fine-tuning dataset size. Note that we only use 50k data in LoRA, and full fine-tuning compares to 500k used in MDT paper 2. The advantages of LoRA include its ability to tackle catastrophic forgetting, as well as its superior performance compared to standard fine-tuning in scenarios with limited data and those that are outside of the usual domain.

\subsection{Analyze of input misleading}

we conducted an experiment to assess the robustness of the proposed method against input distortion. This involved adding Gaussian noise to the input frames of Atari games. Specifically, we set the mean to 0 and experimented with various standard deviation values. The results are detailed in the table below:

\begin{table}[ht]
\centering
\begin{tabular}{c|c|c|c|c}
\hline
                 & Alien  & MsPacman & SpaceInvaders & StarGunner \\ \hline
MDT              & 3.8\%  & 13.2\%   & 8.6\%         & 2.3\%      \\
DT-Mem           & 51.0\% & 69.3\%   & 53.6\%        & 62.2\%     \\
DT-Mem (std=0.5) & 55.3\% & 67.6\%   & 53.0\%        & 57.8\%     \\
DT-Mem (std=1)   & 35.6\% & 56.1\%   & 40.0\%        & 34.6\%     \\
DT-Mem (std=2)   & 25.9\% & 35.6\%   & 30.5\%        & 21.1\%     \\ \hline
\end{tabular}
\end{table}

From the results above, we conclude that the proposed DT-Mem demonstrates greater robustness to noisy inputs compared to the MDT method. This is evident as the DT-Mem consistently outperforms MDT under various levels of Gaussian noise. Notably, the performance with a standard deviation of 0.5 shows minimal difference compared to the no-noise scenario, illustrating DT-Mem's effectiveness in mitigating the impact of varying input distortions.

\color{black}
\section{Memory Module Visualization}

\begin{figure} [htbp]
     \centering
    \includegraphics[width=0.95\textwidth]{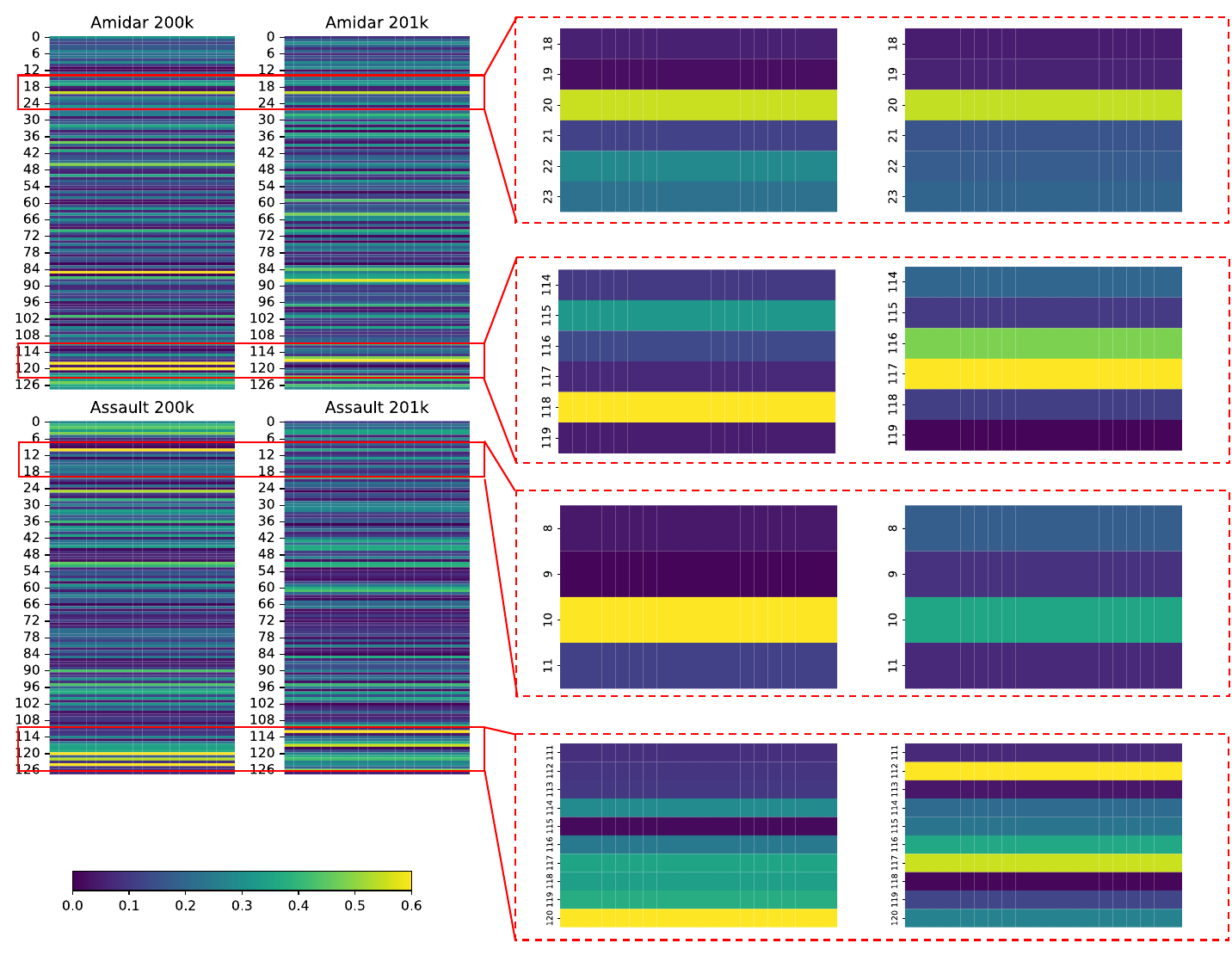}
    \caption{This visualization represents the memory module. In the figure, each row is derived from the mean of a vector that signifies a memory slot. Each depiction calculates the variation between two write operations in a single episode for each memory slot. Lighter shades indicate memory slots that have been actively updated post-write operations. The encircled areas highlight the comparison of active memory slots across different episodes.}
    \label{fig:mem_vis}
\end{figure}

Figure \ref{fig:mem_vis} illustrates the visualization of the memory module. Since memory operations are trained in conjunction with the Transformer module, we randomly select a later training episode to mitigate uncertainties regarding operational parameters. Due to time constraints, we trained on only two games simultaneously. In the revised version of the paper, we intend to provide visualizations for all games. For clearer visualization, we opted for a memory module of a smaller size, containing 128 memory slots.

Let's first discuss how memory modules update within the same game. As observed in the figure, for the Amidar game, the actively updated memory slots concentrate around rows 18, 84, and 117. This pattern is consistent across episodes, albeit with reduced activity. Such a trend indicates that during each training iteration, the Transformer agent tends to overwrite the same memory slot contents. We note a similar observation in the Assault game. Furthermore, we observe that the memory module's activity diminishes in later episodes. For instance, in the Assault game, the active memory slot in row 12 during episode 200k becomes less active by episode 201k. We hypothesize that as training progresses, the accumulated knowledge becomes sufficiently robust for retrieval, reducing the need for updates.

Moving on, when comparing the activity of memory slots across different games, there are intriguing overlaps. For instance, comparing Amidar 200k and Assault 200k reveals that memory slots around row 120 are active in both games. We surmise that this region retains cross-task knowledge shared between games. Additionally, the varying attention across other memory slots demonstrates how these slots assist the agent in decision-making across diverse games.

\section{Limitations}

The first limitation of our work is the sample efficiency of memory fine-tuning. 
The 10\% fine-tuning dataset is still sizeable, and we plan to explore more sample-efficient methods in the future.
We could, for instance, consider a setting with more tasks, each one with less data, so that the inter-task generalization would be even more crucial to its performance.
This work does not propose a control strategy for collecting data on a new task. 
For future work, we plan to investigate online data collection methods, including designing and learning exploration strategies for efficient fine-tuning on new tasks.
Finally, the approach has been intuitively motivated, and it would be valuable to have a theoretical grounding that would show the structural limits of large models and how equipping them with a memory component overcomes them.

\section{Comparison of DT-Mem and Neural Episodic Control (NEC) in Writing and Reading Memory}

\subsection*{Memory Mechanism}
\begin{itemize}
    \item \textbf{NEC:} Utilizes a Differentiable Neural Dictionary (DND) for storing experiences with separate memories for each action, containing state representations (keys) and value function estimates (values).
    \item \textbf{DT-Mem:} Integrates a working memory module within a Transformer framework, focusing on storing, blending, and retrieving information for improving training efficiency and generalization.
\end{itemize}

\subsection*{Writing to Memory}
\begin{itemize}
    \item \textbf{NEC:} Continuously adds new experiences and rapidly updates value function estimates in memory.
    \item \textbf{DT-Mem:} Modifies or replaces existing information in the memory matrix using an attention mechanism to calculate correlations and update memory with the attended weight of the input sequence.
\end{itemize}

\subsection*{Reading from Memory}
\begin{itemize}
    \item \textbf{NEC:} Implements context-based lookups in the DND to retrieve values, outputting a weighted sum based on the similarity between the current state and stored keys.
    \item \textbf{DT-Mem:} Employs content-based addressing for memory retrieval, using attention mechanisms to read from the updated memory and inform decision-making.
\end{itemize}

\subsection*{Distinctive Features and Advantages}
\begin{itemize}
    \item \textbf{NEC:} Designed for rapid assimilation and action upon new experiences with specialized and swift updates for each action.
    \item \textbf{DT-Mem:} Aims to enhance generalization across tasks and reduce catastrophic forgetting by integrating memory with the Transformer's sequential data handling capabilities.
\end{itemize}

\end{document}